\documentclass{article}

\PassOptionsToPackage{numbers, compress}{natbib}

\usepackage[eandd, final]{neurips_2026}

\usepackage[utf8]{inputenc} 
\usepackage[T1]{fontenc}    
\usepackage{hyperref}       
\usepackage{url}            
\usepackage{booktabs}       
\usepackage{amsfonts}       
\usepackage{nicefrac}       
\usepackage{microtype}      
\usepackage{xcolor}         

\usepackage{array}   
\usepackage{adjustbox}      
\usepackage{siunitx}     
\usepackage{tabularx}    
\usepackage{graphicx}    
\usepackage{caption}     
\usepackage{multirow}

\usepackage{amssymb}
\usepackage{arydshln}
\usepackage{pifont}
\usepackage{makecell}    
\usepackage{wrapfig} 
\usepackage{colortbl}
\usepackage[table]{xcolor}   
\newcommand{\cmark}{\ding{51}}  
\newcommand{\xmark}{\ding{55}}  

\definecolor{predcolor}{RGB}{235,245,255}  
\definecolor{gtcolor}{RGB}{255,240,230}    

\usepackage{xcolor}
\usepackage{hhline}   

\newcommand{\promptbox}[1]{%
\vspace{4pt}
\noindent
\fcolorbox{gray!50}{gray!10}{%
\parbox{0.98\linewidth}{%
\vspace{4pt}
\small
#1
\vspace{4pt}
}%
}
\vspace{6pt}
}

\title{Towards Characterizing Scientific Image Utility and Upgradability}

\author{
Wenzhe Li\textsuperscript{1,2},
Qihang Yan\textsuperscript{2},
Liang Chen\textsuperscript{2,3}, 
Junying Wang\textsuperscript{2}, 
Yijing Guo \textsuperscript{2},\\ 
\textbf{
Farong Wen \textsuperscript{2}, 
Chunyi Li \textsuperscript{2}, 
Zicheng Zhang \textsuperscript{2\dag}, 
Guangtao Zhai \textsuperscript{2,3\dag}
}\\
\textsuperscript{1}Shanghai Research Institute for Intelligent Autonomous Systems, TongJi University\\
\textsuperscript{2}Shanghai Artificial Intelligence Laboratory 
\textsuperscript{3}Shanghai Jiao Tong University\\
\textsuperscript{\dag}Corresponding authors\\
\
}

\begin{document}

\maketitle

\begin{abstract}

Scientific images function as critical evidence in research communication, yet their integrity faces unprecedented threats from AI-generated content that introduces subtle but consequential errors. Existing evaluation paradigms prove inadequate: perceptual quality metrics poorly correlate with scientific validity, while language models lack domain-specific verification capabilities.
To address this gap, we propose the \textbf{S}cientific \textbf{I}mage \textbf{U}tility and \textbf{U}pgradability \textbf{A}ssessment (\textbf{SIU$^2$A}) framework, which introduces two complementary dimensions for scientific image evaluation. \textbf{Utility} encompasses \textit{error detection} (identifying scientific inaccuracies) and \textit{correction feasibility} (assessing whether errors can be reliably repaired). \textbf{Upgradability} measures the quality of correction.
We categorize scientific image corruption into four fundamental types: Detail Distortion, Incompleteness, False Content, and Entity Confusion. Based on this taxonomy, we construct SIU$^2$A-Benchmark, a dataset with expert annotations for error identification and repair. The framework implements a two-stage evaluation protocol: the \textit{Utility} stage evaluates error detection capability and repair instruction generation, while the \textit{Upgradability} stage assesses whether corrections faithfully restore scientific validity without compromising existing accurate information.
Experiments reveal that current multimodal systems exhibit significant limitations in both scientific error assessment and faithful correction, exposing a fundamental gap between visual perception and scientific usability.

\textbf{Keywords: Scientific Image Utility, Multimodal Scientific Reasoning, AI-Generated Image Evaluation
}
\end{abstract}

\begin{figure}[htbp]
    \vspace{-1.5em}
    \centering
    \includegraphics[width=0.8\linewidth]{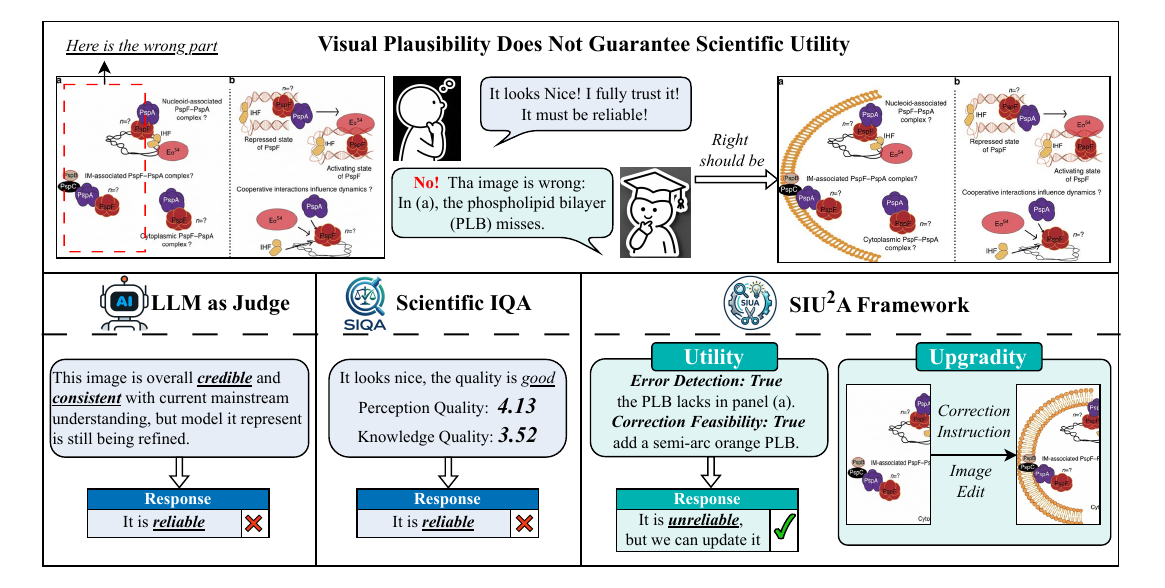}
    \caption{Visual plausibility is not scientific validity. Traditional assessors (LLM/S-IQA) are fooled by stealthy errors, but SIU$^2$A grounded in utility and upgradability is built to detect and correct them.}
    \label{fig:motivation}
    \vspace{-1.5em}
\end{figure}

\section{Introduction}

Scientific images serve not merely as illustrations, but as primary carriers of experimental evidence and scientific claims. Their value lies in their functional role in conveying scientific evidence rather than plausible appearance. Even subtle defects, such as missing scale bars, mislabeled axes, or distorted data curves, can invalidate conclusions. The widespread adoption of image editing tools and generative AI has made such corruptions increasingly prevalent, underscoring the critical need for vigilance in scientific communication.

Existing approaches remain limited as shown in Fig.~\ref{fig:motivation}. Traditional image quality assessment (IQA) methods focus on low-level distortions such as noise or blur, which are weakly correlated with scientific validity \cite{CNN-IQA1, CNN-IQA2, Q-Align, Q-Eval,iqa1}. Multimodal models, while capable of describing image content, lack explicit mechanisms to assess scientific correctness or reason about domain-specific plausibility \cite{zhang2025large, A-Bench, iqa2, iqa3, iqa4}. More importantly, current methods are largely diagnostic: they can identify anomalies but do not provide a pathway for correcting them, limiting their usefulness in real scientific workflows where many errors are accidental and potentially recoverable.

To address this gap, the \textbf{S}cientific \textbf{I}mage \textbf{U}tility and \textbf{U}pgradability Assessment (SIU$^2$A) framework is proposed. Scientific image utility is defined as a joint function of two properties: (i) \textbf{utility}, which evaluates the two-stage capability: first, to reliably detect, precisely localize, and structurally describe scientific errors in images; and second, to generate actionable correction instructions that enable accurate restoration of corrupted content. 
(ii) \textbf{upgradability}, which measures whether the image can be faithfully restored through corrective actions while preserving existing valid scientific content and maintaining consistency of underlying scientific knowledge. SIU$^2$A thus shifts image evaluation from passive assessment toward actionable and consistency-preserving restoration.

Formally, SIU$^2$A is instantiated through four types of scientific image corruption: Detail Distortion, Incompleteness, False Content, and Entity Confusion. Each type is simulated using a corresponding primitive corruption operation (modify, remove, add, and swap), which enables controlled construction of realistic scientific image failures.
Based on this formulation, SIU$^2$A-Benchmark is constructed as a curated dataset of scientific images spanning multiple disciplines as shown in Table~\ref{tab:data_compare}, including biology, chemistry, and physics. Each sample is paired with controlled corruptions that simulate realistic scientific errors while preserving ground-truth recoverability. Expert annotations further determine whether each case is diagnosable and upgradable, enabling systematic evaluation of both error understanding and restoration feasibility.
A two-stage evaluation protocol is further defined, aligned with real scientific workflows: error detection, structured diagnosis, and image restoration. This pipeline moves beyond static evaluation toward an interactive loop of error identification and correction.

The main contributions are summarized as follows:
\begin{itemize}
    \item (1) SIU$^2$A is proposed as a unified framework that models scientific image utility as a combination of structured error diagnosability (detection, localization, and instruction generation) and consistency-preserving upgradability (faithful restoration of scientific validity without compromising existing correct information).

    \item (2) SIU$^2$A-Benchmark is constructed as a structured dataset that enables systematic evaluation of scientific image corruption under controlled operations, with annotations for both utility and upgradability.

    \item (3) Extensive experiments demonstrate that current multimodal understanding and image editing models fail to reliably perform structured scientific error localization and consistency-preserving restoration, revealing a fundamental gap in utility-aware visual reasoning.
\end{itemize}


\begin{table}[htbp]
\vspace{-1.5em}
\centering
\caption{Comparison of existing multimodal scientific reasoning and image editing benchmarks. SIU$^2$A uniquely supports both error-aware diagnosis and image restoration within a unified framework.}
\label{tab:data_compare}
\begin{tabular}{
lccccc
}
\hline
Dataset & Domain & Reason & Edit & \makecell{Error-Aware} & Task \\
\hline
GeoTrust & Geometric & \cmark & \xmark & \xmark & VQA \\
ChemVLM & Chemical  & \cmark & \xmark & \xmark & VQA \\
ScienceQA & Science & \cmark & \xmark & \xmark & VQA \\
I2EBench & Natural & \xmark & \cmark & \xmark & Edit \\
Ebench & Natural & \xmark & \cmark & \xmark & Edit \\
GRADE & Science & \xmark & \cmark & \xmark & Edit \\
\hline
\textbf{SIU$^2$A (Ours)} & Science & \cmark & \cmark & \cmark & VQA \& Edit \\
\hline
\end{tabular}
\vspace{-1.5em}
\end{table}

\section{Related Work}

\subsection{MLLMs for Scientific Understanding}

Recent advances in multimodal large language models (MLLMs) have established a unified paradigm for scientific knowledge understanding and task execution, enabling a shift from isolated capabilities toward system-level scientific intelligence.
Scientific foundation models such as Galactica \cite{taylor2022galactica}, Kosmos-2.5 \cite{lv2023kosmos}, ChemDFM \cite{zhao2024chemdfm}, and InstructMol \cite{cao2025instructmol} demonstrate strong cross-modal scientific representation and reasoning abilities in domains such as molecular modeling and scientific inference.
At a higher level, recent efforts like DeepScientist~\cite{weng2026deepscientist} extend MLLMs toward scientific agents for workflow automation, including literature analysis, hypothesis generation, and scientific assistance. Meanwhile, benchmarks such as SciBench \cite{wang2024scibenc} evaluate structured scientific reasoning capabilities.
Despite these advances, existing methods remain largely focused on semantic and textual reasoning, with limited explicit modeling of errors in scientific visual data.
As AIGC systems increasingly generate and modify scientific images, visually plausible but scientifically incorrect artifacts become more frequent and harder to detect.
This highlights the need for systematic evaluation of scientific image inconsistencies and their potential for correction, which motivates SIU$^2$A-Utility for assessing error identification and correction reasoning in MLLMs.

\subsection{Image Editing Models and Evaluation}

Image editing has evolved from diffusion-based generation frameworks to instruction-driven and multimodal editing systems, significantly improving controllability and visual quality. Stable Diffusion \cite{rombach2021highresolution} and InstructPix2Pix \cite{brooks2022instructpix2pix} provide foundational capabilities, while recent multimodal editing systems (e.g., GPT-Image~\cite{openai2025gptimage15,openai_gptimage2}, Qwen-Image~\cite{wu2025qwenimagetechnicalreport}, Gemini-Image~\cite{gemini3proimage,google2025gemini25image}) further enhance natural image editing performance, though their effectiveness in scientific domains remains underexplored.
Evaluation methods include pixel-level metrics (PSNR, SSIM), perceptual metrics (LPIPS, FID), and recent benchmarks such as I2I-Bench \cite{wang2025i2i} and LMM4Edit \cite{xu2025lmm4edit}, which evaluate instruction following and multi-dimensional editing quality. GRADE \cite{liu2026grade} further introduces scientific image editing evaluation, but remains focused on instruction adherence and knowledge consistency, without explicitly modeling scientific correctness or error correction.

Overall, existing methods emphasize visual quality and instruction alignment, but lack systematic modeling of scientific error correction and validity restoration.
This motivates viewing scientific image editing as a structured problem involving error identification, correction reasoning, and faithful execution of corrective instructions to restore scientific validity.

\begin{figure}[t]
    \centering
    \includegraphics[width=1\linewidth]{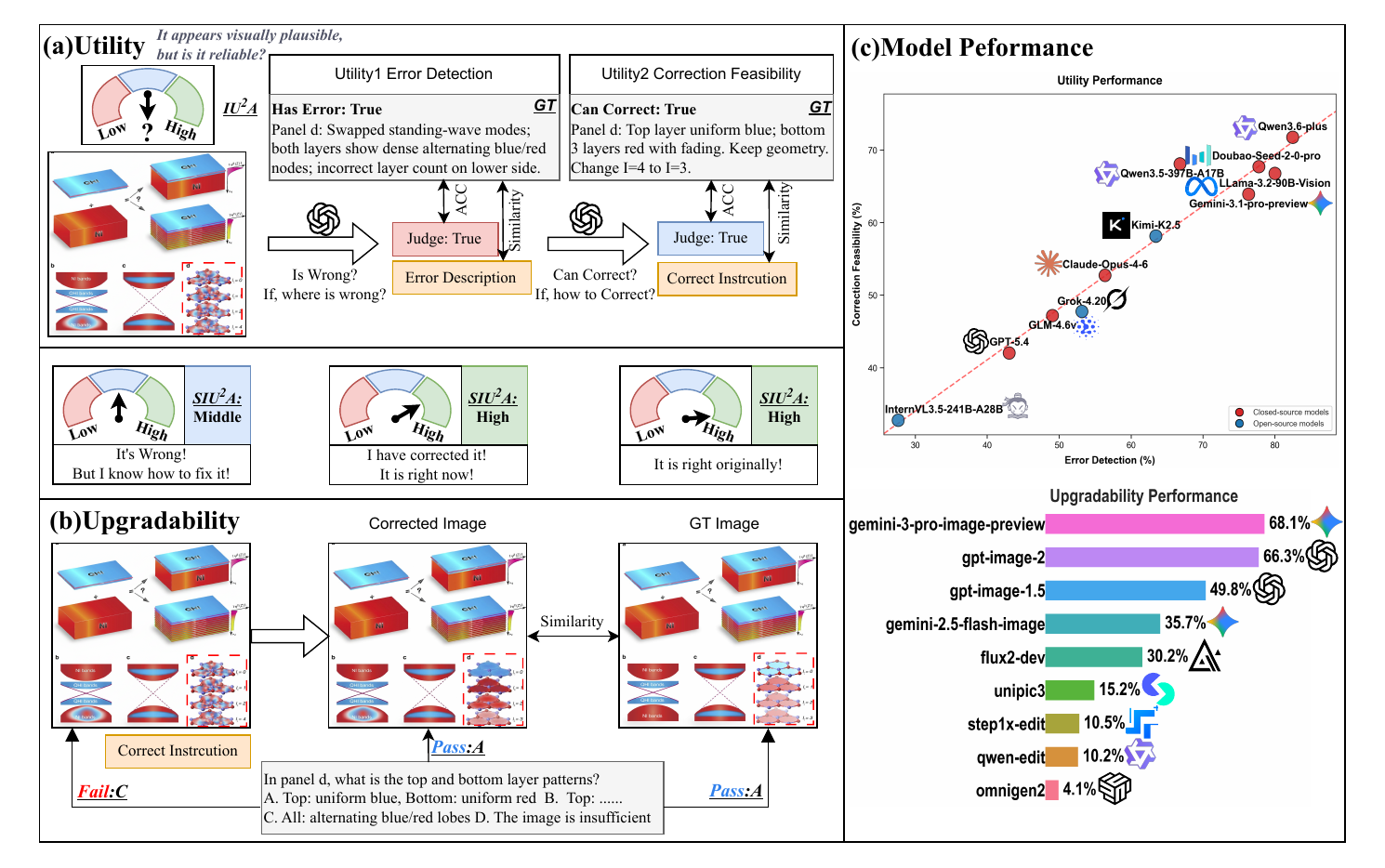}
    \caption{The SIU$^2$A framework for scientific image assessment: (a) utility via error detection and correction feasibility, (b) upgradability through a diagnosis-to-correction pipeline, and (c) comparative model performance.}
    \label{fig:spotlight}
    \vspace{-1.5em}
\end{figure}

\section{SIU$^2$A Framework}

The utility of a scientific image is fundamentally determined by whether its underlying scientific validity can be reliably preserved or restored. An image remains useful if its errors are both detectable and correctable through grounded reasoning. Otherwise, it may lead to unreliable scientific conclusions although it looks plausible. 
Based on this principle, we introduce the \textbf{Scientific Image Utility and Upgradability Assessment (SIU$^2$A)} framework, which jointly models scientific image diagnosis, correction, and edit validation under realistic corruption conditions. As shown in Fig.~\ref{fig:spotlight}, SIU$^2$A explicitly evaluates whether scientific validity can be identified, reasoned about, and faithfully restored through structured intervention, thereby shifting image evaluation from passive assessment to actionable restoration.

\subsection{SIU$^2$A Definition}

\paragraph{Scientific Images Failure Summary}
We summarize the common failure modes in scientific images into four structurally distinct categories:
\textbf{(i) Detail Distortion}, where low-level visual signals are corrupted and local structures become inconsistent, leading to unreliable perception of fine-grained scientific patterns such as curves, boundaries, or measurement markings;
\textbf{(ii) Incompleteness}, where essential scientific elements are missing, resulting in partial observations that prevent correct inference of the underlying scientific phenomenon;
\textbf{(iii) False Content}, where spurious or scientifically invalid elements are introduced, potentially inducing incorrect evidence interpretation and misleading causal or quantitative reasoning;
\textbf{(iv) Entity Confusion}, where semantic mappings between visual regions and scientific entities are misassigned or swapped, causing incorrect association between visual evidence and its scientific meaning, and thus breaking the consistency of downstream reasoning.
These error categories are instantiated through four primitive entity-level corruption operations: \textit{Modify}, \textit{Remove}, \textit{Add}, and \textit{Swap}, providing a controllable mechanism to simulate realistic scientific failure cases.

\paragraph{Evaluation along Utility and Upgradability}
Given the above error space, we define scientific image usability along two complementary dimensions that jointly characterize the full lifecycle of error understanding and correction.
\textbf{Utility} quantifies a model's ability to ground and structure scientific errors in images through two sequential capabilities:
\textit{Error Detection:} The model first identifies whether scientific inconsistencies exist and, when present, provides accurate, grounded error descriptions.
\textit{Correction Feasibility:} Conditioned on detected errors, the model assesses their repairability and, if feasible, generates grounded correction instructions that specify how scientific validity can be restored.
\textbf{Upgradability} measures the ability to realize corrections from given instructions. Formally, given an image and its correction instruction, the model must generate a restored image that resolves the original error while preserving valid scientific structures and semantic consistency. This dimension therefore evaluates whether scientific validity can be effectively recovered through intervention rather than only described.
Together, Utility and Upgradability form a unified evaluation of scientific image reliability under corruption, where the former captures error understanding and specification, and the latter captures execution of error recovery.

\begin{figure}[t]
    \centering
    \includegraphics[width=1\linewidth]{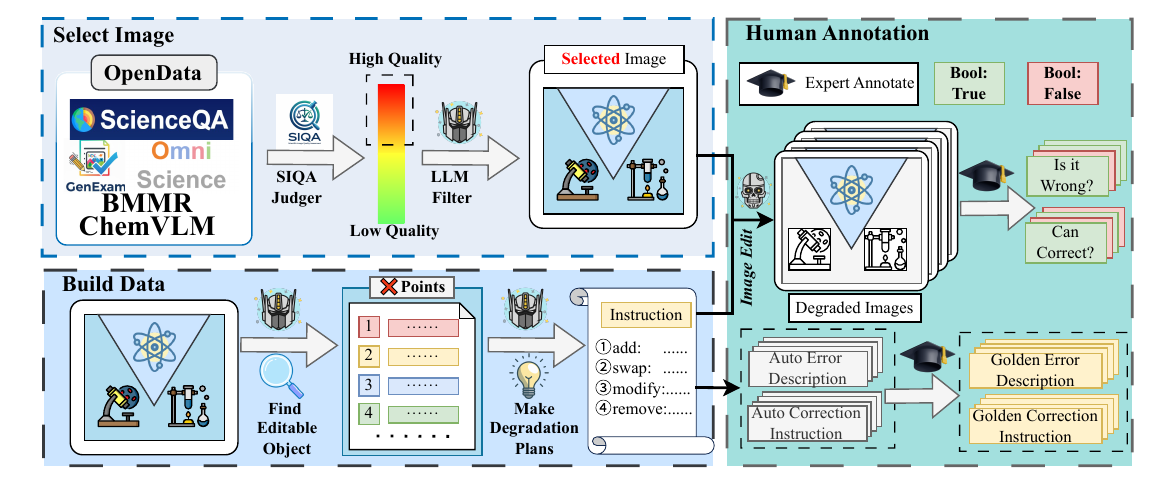}
    \caption{Pipeline for constructing the SIU$^2$A-Benchmark dataset, including high-quality scientific image filtering, controlled degradation generation, and expert annotation.}
    \label{fig:build_dataset}
    \vspace{-1.5em}
\end{figure}

\subsection{SIU$^2$A-Benchmark Construction.}

\paragraph{Base Image Collection}
To support the above formulation, we construct \textit{SIU$^2$A-Benchmark}, a dataset that jointly evaluates diagnosis, instruction generation, and editing under controlled scientific corruptions as shown in Fig.~\ref{fig:build_dataset}. Scientific images are collected from multiple public sources, including ScienceQA~\cite{lu2022learn}, GenExam~\cite{wang2025genexam}, BMMR~\cite{xi2025bmmrlargescalebilingualmultimodal}, ChemVLM~\cite{li2025chemvlm}, and OmniScience~\cite{tao2026omniscience}. To ensure both quality and editability, a two-stage filtering process is applied: a scientific image quality assessment model (SIQA~\cite{li2026siqa}) first removes low-quality or ambiguous samples, followed by MLLMs verification to retain images containing explicit and structurally editable scientific elements such as charts, annotations, and labeled components. This yields 600 high-quality base images.

\paragraph{Structured Perturbation}
Controlled perturbations are then generated for each image according to the four SIU$^2$A error types. Editable regions are identified automatically, and candidate perturbations are proposed to simulate realistic scientific errors (e.g., missing annotations, incorrect values, or entity mismatches). One perturbation per error type is selected and applied using an image editing model, resulting in approximately 2,100 corrupted–clean image pairs that cover diverse scientific failure modes.

\begin{figure}[t]
    \centering
    \includegraphics[width=1\linewidth]{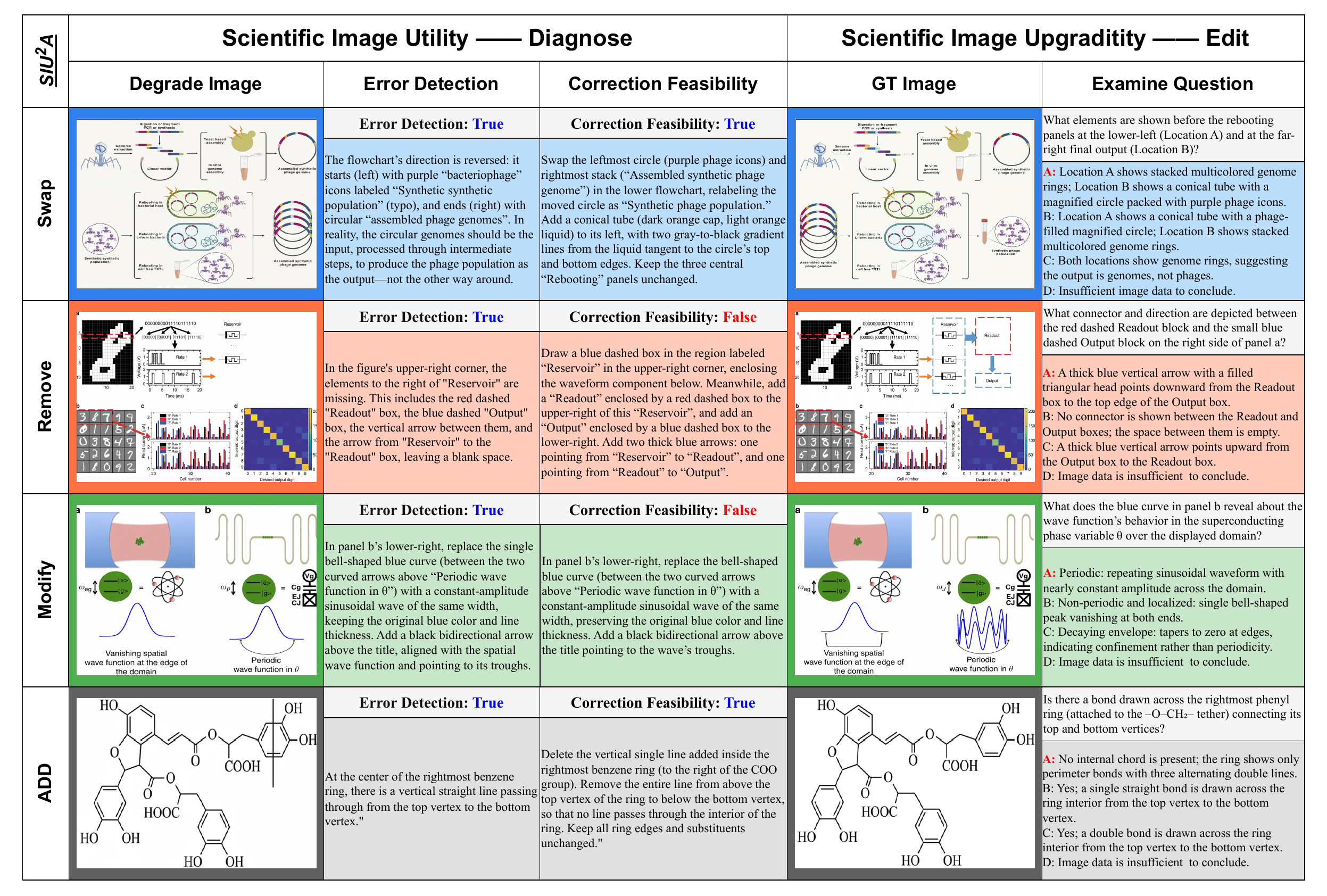}
    \caption{Overview of the SIU$^2$A data structure. Each instance contains a ground-truth image, a corrupted image, error detection and correction feasibility labels, structured error descriptions, correction instructions, and a corresponding scientific QA pair.}
    \label{fig:datacase}
    \vspace{-2em}
\end{figure}

\paragraph{Expert Annotation}
To ensure scientific validity and annotation consistency, an expert-in-the-loop protocol is adopted. The dataset is first partitioned into \textit{Advanced} and \textit{Simple} subsets using LLM-assisted difficulty assessment, where Advanced samples require multi-step reasoning or cross-entity understanding. 
A total of 600 Advanced samples are selected for full expert annotation, including perturbation validation, utility labeling(binary assessment of error detectability and correction feasibility), and the error descriptions and correction instructions. 
The remaining 1,800 Simple samples undergo lightweight expert verification, with invalid or infeasible cases removed. The final dataset consists of 400 Advanced and 1,200 Simple instances, annotated by ten domain experts across multiple scientific disciplines under a unified guideline. The details can refer to Fig.~\ref{fig:datacase}.

\subsection{Evaluation Protocol.}
We define a unified evaluation framework that jointly measures \textit{decision correctness} and \textit{semantic fidelity} across SIU$^2$A-Diagnose and SIU$^2$A-Edit. Metrics are organized under two complementary paradigms: (i) \textbf{no-reference evaluation}, which assesses outputs independently, and (ii) \textbf{full-reference evaluation}, which measures consistency with ground truth.
For \textbf{Utility}, (1) \textit{Diagnosis Accuracy} evaluates whether the model correctly predicts image diagnosability as a binary classification task based on expert-annotated labels, and (2) \textit{Text Consistency} measures semantic alignment between generated error descriptions and expert annotations using BERTScore and LLM-as-a-Judge.
For \textbf{Upgradability}, (1) \textit{Problem Resolution Rate} evaluates whether the edited image restores scientific utility without access to the original image, implemented via QA-based probes, and (2) \textit{Scientific Semantic Consistency} measures whether the edited result preserves semantic consistency with the original image using structured LLM-based comparison.
Overall, this framework explicitly disentangles functional correctness (task completion) from semantic faithfulness (scientific validity preservation), enabling a comprehensive evaluation of both error understanding and correction quality.

\begin{table}[t]
\centering
\caption{Performance comparison on Utility tasks. Models are evaluated in a two-stage pipeline where they first detect and localize scientific errors in images, and then generate correction instructions conditioned on the predicted errors for downstream restoration. Best and second-best results are highlighted in \textbf{bold} and \underline{underlined}, respectively.}
\label{tab:diagnosis_repair}

\begin{tabular}{
l|
*{5}{>{\centering\arraybackslash}m{0.55cm}}
|
*{5}{>{\centering\arraybackslash}m{0.55cm}}
}

\hline
\hline

\textbf{Error Detection} & \multicolumn{5}{c|}{\textbf{Error-ACC}} 
& \multicolumn{5}{c}{\textbf{Error-Text}} \\
\hline

\textbf{Model} & \textbf{Add} & \textbf{Swap} & \textbf{Mod.} & \textbf{Rem.} & \textbf{All}
& \textbf{Add} & \textbf{Swap} & \textbf{Mod.} & \textbf{Rem.} & \textbf{All} \\
\hline
\multicolumn{11}{l}{\textit{Closed-source}} \\
\hdashline

Qwen3.6-plus~\cite{qwen36plus} & \textbf{84.5} & \textbf{83.8} & \textbf{83.5} & \underline{78.0} & \textbf{82.5} & \underline{24.9} & \textbf{23.9} & \underline{23.4} & 14.3 & \underline{21.8} \\
Doubao-Seed-2.0-Pro~\cite{bytedance2026seed2} & 79.4 & 78.4 & \underline{79.4} & 73.7 & 77.8 & 24.3 & 21.8 & 23.5 & \underline{15.3} & 21.4 \\
Gemini-3.1-Pro~\cite{google2026gemini31propreview} & 79.0 & 79.4 & 74.3 & 72.6 & 76.3 & \textbf{26.1} & 21.9 & \textbf{23.8} & \textbf{16.2} & \textbf{22.2} \\
Claude-Opus-4.6~\cite{anthropic2026opus46} & 55.5 & 58.8 & 55.9 & 55.5 & 56.4 & 9.17 & 13.0 & 13.7 & 5.63 & 10.4 \\
Grok-4.20~\cite{grok4_reasoning_2026} & 52.6 & 59.3 & 53.0 & 47.8 & 53.2 & 13.6 & 13.7 & 15.9 & 7.37 & 12.7 \\
GPT-5.4~\cite{openai2026gpt54} & 43.8 & 43.6 & 44.4 & 40.4 & 43.1 & 5.62 & 8.64 & 8.21 & 3.25 & 6.46 \\

\hline
\multicolumn{11}{l}{\textit{Open-source}} \\
\hdashline

Qwen3.5-397B~\cite{qwen3.5} & 75.6 & 77.5 & 77.0 & 71.4 & 76.3 & 20.8 & \underline{22.4} & 23.3 & 13.6 & 23.4 \\
Kimi-K2.5~\cite{team2026kimi} & 62.4 & 65.5 & 62.4 & 63.7 & 63.4 & 15.2 & 13.9 & 18.6 & 11.9 & 15.0 \\
LLaMa-3.2-90B-Vision~\cite{llama32} & \underline{80.0} & \underline{82.3} & 79.1 & \textbf{79.0} & 79.4 & 3.71 & 4.82 & 4.07 & 2.51 & 3.28 \\
GLM-4.6V~\cite{GLM46v} & 46.7 & 51.3 & 50.6 & 47.8 & 49.1 & 6.93 & 7.83 & 8.87 & 3.22 & 6.77 \\
InternVL3.5-241B~\cite{wang2025internvl3} & 27.6 & 26.1 & 28.3 & 24.8 & 31.4 & 1.50 & 1.14 & 1.94 & 0.74 & 2.17 \\

\hline
\hline

\textbf{Correction Feasibility} & \multicolumn{5}{c|}{\textbf{Correction-ACC}} 
& \multicolumn{5}{c}{\textbf{Correction-Text}} \\
\hline

\textbf{Model} & \textbf{Add} & \textbf{Swap} & \textbf{Mod.} & \textbf{Rem.} & \textbf{All} & \textbf{Add} & \textbf{Swap} & \textbf{Mod.} & \textbf{Rem.} & \textbf{All} \\

\hline
\multicolumn{11}{l}{\textit{Closed-source}} \\
\hdashline

Qwen3.6-plus~\cite{qwen36plus} & \textbf{77.3} & \textbf{73.1} & \textbf{73.6} & \textbf{62.6} & 71.8 & 20.8 & \underline{17.2} & 22.3 & 11.2 & 18.2 \\
Doubao-Seed-2.0-Pro~\cite{bytedance2026seed2} & \underline{72.3} & 69.5 & \underline{70.5} & 58.2 & 67.7 & \underline{23.5} & \textbf{18.7} & \textbf{24.1} & \textbf{13.9} & \textbf{20.4} \\
Gemini-3.1-Pro~\cite{google2026gemini31propreview} & 69.6 & 64.1 & 64.2 & 57.4 & 63.9 & \textbf{24.1} & 14.8 & 22.3 &\underline{ 12.2} & \underline{18.8} \\
Claude-Opus-4.6~\cite{anthropic2026opus46} & 52.4 & 53.2 & 53.0 & 52.3 & 52.7 & 7.51 & 10.0 & 12.7 & 4.86 & 8.92 \\
Grok-4.20~\cite{grok4_reasoning_2026} & 49.8 & 48.3 & 48.4 & 44.4 & 47.8 & 13.1 & 10.4 & 14.8 & 5.27 & 11.2 \\
GPT-5.4~\cite{openai2026gpt54} & 40.4 & 41.1 & 44.6 & 42.1 & 42.0 & 5.68 & 7.04 & 8.72 & 2.45 & 6.12 \\

\hline
\multicolumn{11}{l}{\textit{Open-source}} \\
\hdashline

Qwen3.5-397B~\cite{qwen3.5} & 68.1 & \underline{71.3} & 69.0 & \underline{61.0} & \underline{70.8} & 18.1 & 19.3 & \underline{22.6} & 11.1 & 18.1 \\
Kimi-K2.5~\cite{team2026kimi} & 57.2 & 60.5 & 59.6 & 55.4 & 58.2 & 14.5 & 11.6 & 18.3 & 10.2 & 13.8 \\
LLaMa-3.2-90B-Vision~\cite{llama32} & 66.8 & 71.5 & 67.8 & \underline{61.0} & 66.4 & 4.53 & 4.17 & 6.20 & 2.84 & 4.63 \\
GLM-4.6V~\cite{GLM46v} & 43.8 & 47.6 & 52.8 & 44.6 & 47.2 & 7.00 & 4.49 & 8.66 & 2.14 & 5.77 \\
InternVL3.5-241B~\cite{wang2025internvl3} & 32.8 & 28.0 & 33.7 & 36.7 & 33.1 & 1.91 & 1.32 & 2.75 & 0.86 & 2.60 \\

\hline
\hline
\end{tabular}
\vspace{-1.5em}
\end{table}

\section{Experiments}

\subsection{Experimental Setup}

We evaluate state-of-the-art MLLMs on the SIU$^2$A-Utility and SIU$^2$A-Upgradability tasks. For SIU$^2$A-Utility, we assess a set of recent proprietary and open-source MLLMs, including 
Qwen3.6-Plus~\cite{qwen36plus}, Doubao-Seed-2-Pro~\cite{bytedance2026seed2}, Gemini-3.1-Pro-Preview~\cite{google2026gemini31propreview}, Kimi-K2.5~\cite{team2026kimi}, Claude-Opus-4.6~\cite{anthropic2026opus46}, GLM-4.6V~\cite{GLM46v}, Grok-4.20~\cite{grok4_reasoning_2026}, GPT-5.4~\cite{openai2026gpt54},LLaMa-3.2-90B~\cite{llama32}, Qwen3.5~\cite{qwen3.5}, and InternVL3.5~\cite{wang2025internvl3}.
All MLLMs are prompted to output in a standardized JSON format, explicitly indicating whether they can perform the diagnosis task and providing the corresponding error description text when applicable.
For SIU$^2$A-Upgradability, we evaluate existing image editing models including 
GPT-Image-1.5~\cite{openai2025gptimage15},GPT-Image-2~\cite{openai_gptimage2}, Gemini-2.5-Image~\cite{google2025gemini25image}, Gemini-3-Image-Pro~\cite{gemini3proimage}, Flux2-Dev~\cite{flux-2-2025}, Step1X-Edit~\cite{liu2025step1x-edit}, UniPic3~\cite{wei2026skyworkunipic30unified}, Qwen-Image-Edit-2511~\cite{wu2025qwenimagetechnicalreport}, and OmniGen2~\cite{wu2025omnigen2}
by directly feeding them the degraded image along with the editing instruction to produce the repaired output, without requiring intermediate diagnostic judgments.

\subsection{Findings of Utility Performance Analysis}

\paragraph{Detection is largely solved, but precise error description and correction instruction remain fundamental challenges.}
Table~\ref{tab:diagnosis_repair} reveals a cascading degradation in multimodal scientific reasoning under the SIU$^2$A framework. 
Across all models, Error-ACC (error detection accuracy) consistently outperforms both Error-Text (textual description quality) and downstream correction metrics, indicating that identifying visual anomalies is significantly easier than articulating them as structured scientific statements.
For instance, Qwen3.6-plus achieves strong Error-ACC (82.5\%), but its Error-Text score drops to 21.8\%, while further declining to 18.2\% in Correction-Text, highlighting a progressive degradation from detection to structured description and downstream correction.
Across error categories, a stable ordering is observed, with \textit{Remove-type} errors consistently being the most challenging. This suggests that Remove-type errors are inherently more challenging, as they require reasoning over structural completeness and missing scientific entities.
Overall, the results reveal a cascading failure in SIU$^2$A reasoning. 
While models can reliably detect errors, they fail to produce high-quality textual descriptions that accurately ground visual evidence into structured scientific statements.
Since Correction Feasibility is conditioned on both the image and the generated error description, deficiencies in this intermediate representation propagate to the second stage, leading to degraded feasibility judgments and correction instructions.
These results identify a key bottleneck in SIU$^2$A reasoning: the construction of accurate intermediate textual representations, a finding further validated by ablation studies showing that improving textual grounding consistently enhances correction performance.

\begin{figure}[htbp]
    \centering
    \includegraphics[width=0.6\linewidth]{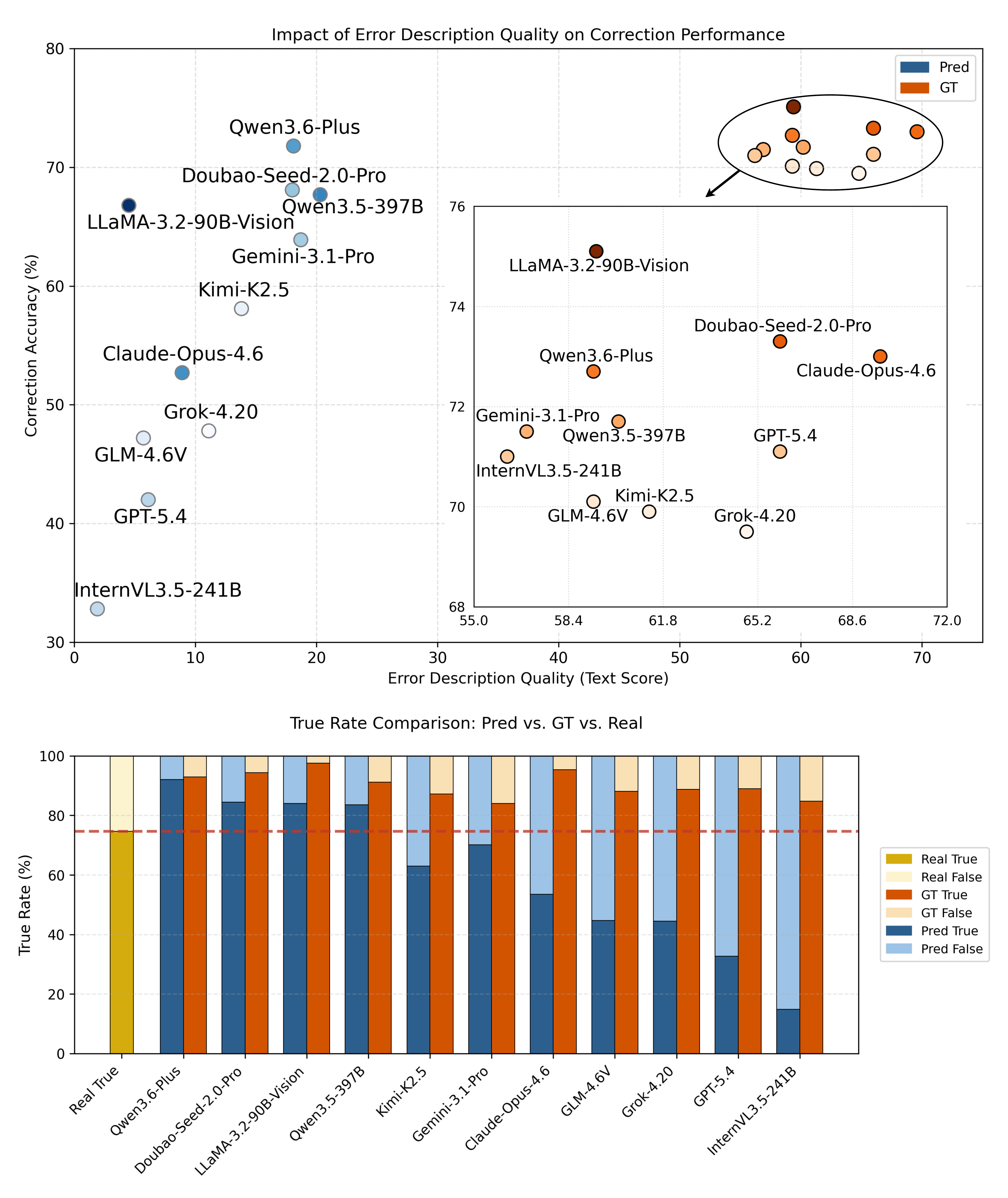} 
    \caption{Ablation study on the impact of error description quality on correction performance. We compare ground-truth (GT) and model-predicted (Pred) error descriptions.}
    \label{fig:ablation1}
    \vspace{-1.5em}
\end{figure}

\paragraph{Decisive Impact of Error Description Quality on Correction Feasibility Performance}
The Fig.~\ref{fig:ablation1} shows that correction feasibility performance is strongly influenced by the quality of generated error descriptions, with clear differences across model capabilities.
For high-capability models such as Qwen3.6-plus, correction feasibility accuracy remains stable between ground-truth (GT) and predicted error descriptions, indicating that their correction judgments are insensitive to variations in error description quality.
In contrast, lower-capability models like GPT-5.4 exhibit sensitivity to error description quality: accuracy improves substantially when provided with GT. This dependence reveals that their limitations stem primarily from inaccurate error descriptions rather than inherent correction reasoning deficiencies. 
This tendency about description quality is further reflected in the convergence of text score under GT, where all models reach a similar higher level, reducing performance differences in the correction stage. 
Furthermore, we uncover how model prediction biases can artificially inflate accuracy metrics. 
LLaMA-3.2-90B-Vision exhibits an absolute bias toward predicting ``True''. Under predicted error descriptions, despite extremely poor 4.5 text similarity , it achieves 65.7 accuracy.
Under ground-truth error, it predicts ``True'' for nearly all cases with relatively moderate 50 text similarity , resulting in an accuracy of 75.1 that closely approximates the true positive ratio in the dataset. 
This indicates complete failure of LLaMA as a Judge, as it uniformly defaults to the majority class without meaningful discrimination. 
In contrast, InternVL3.5-241B demonstrates an opposite bias toward ``False'' predictions under predicted conditions, with accuracy 33.6 approximately equal to $1 - \text{True rate}$.
However, as text quality improves under GT error descriptions, the model's true positive prediction rate normalizes.
Consequently, the accuracy rises to 71.0, demonstrating its latent capability, revealing the model's underlying capability when guided by accurate error 
Under comparable error description quality, such differences are largely driven by prediction bias rather than reasoning capability.
Overall, these results identify error description generation as the primary bottleneck in SIU$^2$A, limiting the effective utilization of downstream repair models.

\subsection{Findings of Upgradability Performance Analysis}
\begin{table}[t]
\centering
\caption{Upgradability results under ground-truth (GT) correction instructions. All models are evaluated using expert-annotated correction instructions to measure upper-bound editing performance under idealized guidance. \textbf{Best} and \underline{second-best} results are highlighted in bold and underlined, respectively.}
\label{tab:edit_only_gt_only}

\begin{tabular}{
l
|
*{5}{>{\centering\arraybackslash}m{0.55cm}}
|
*{5}{>{\centering\arraybackslash}m{0.55cm}}
}
\hline
\hline
\textbf{Upgradability} & \multicolumn{5}{c|}{\textbf{Similarity (Full)}}
& \multicolumn{5}{c}{\textbf{Question-ACC (None)}} \\

\hline

\textbf{Model}
 & \textbf{Add} & \textbf{Swap} & \textbf{Mod.} & \textbf{Rem.} & \textbf{All}
& \textbf{Add} & \textbf{Swap} & \textbf{Mod.} & \textbf{Rem.} & \textbf{All} \\

\hline
\multicolumn{11}{l}{\textit{Closed-source}} \\
\hdashline

Gemini-3-Image-Pro~\cite{gemini3proimage}
& \textbf{76.8} & \textbf{76.4} & \textbf{86.0} & \textbf{87.9} & \textbf{81.6}
& \underline{73.9} & 62.2 & 67.7 & 67.1 & 68.1 \\

GPT-Image-2~\cite{openai_gptimage2}
& \underline{80.2} & \underline{66.7} & \underline{82.5} & \underline{83.5} & \underline{78.6}
& 63.2 & \underline{65.2} & 67.6 & 69.6 & 66.3 \\

GPT-Image-1.5~\cite{openai2025gptimage15}
& 54.9 & 45.0 & 46.7 & 51.8 & 49.8
& \textbf{81.4} & \textbf{67.8} & \textbf{76.0} & \textbf{80.4} & \textbf{76.7} \\

Gemini-2.5-Flash~\cite{google2025gemini25image}
& 36.8 & 34.8 & 29.5 & 42.2 & 35.7
& 70.8 & 65.6 & \underline{70.2} & \underline{75.8} & \underline{70.6} \\

\hline
\multicolumn{11}{l}{\textit{Open-source}} \\
\hdashline

Flux2-Dev~\cite{flux-2-2025}
& 39.4 & 26.5 & 26.0 & 27.2 & 30.2
& 64.6 & 56.7 & 65.4 & 66.3 & 63.4 \\

UniPic3~\cite{wei2026skyworkunipic30unified}
& 19.7 & 16.1 & 12.3 & 12.2 & 15.2
& 69.9 & 46.7 & 59.6 & 51.1 & 57.6 \\

Qwen-Edit~\cite{wu2025qwenimagetechnicalreport}
& 16.7 & 10.6 & 6.59 & 7.40 & 10.5
& 56.6 & 53.3 & 53.9 & 52.2 & 54.1 \\

Step1X-Edit~\cite{liu2025step1x-edit}
& 12.8 & 8.67 & 11.1 & 7.41 & 10.2
& 65.5 & 64.4 & 58.7 & 48.9 & 59.7 \\

OmniGen2~\cite{wu2025omnigen2}
& 6.23 & 3.61 & 3.91 & 2.23 & 4.11
& 32.7 & 34.4 & 36.5 & 25.0 & 32.3 \\

\hline
\hline

\end{tabular}
\vspace{-1.5em}
\end{table}

\paragraph{Fundamental Challenge in Upgradability: Performance Disparity and Non-Saturation}
Even under ideal correction conditions using ground-truth expert annotations, current image editing models exhibit substantial performance disparities and remain far from saturation in both scientific correctness and semantic consistency. As shown in Table~\ref{tab:edit_only_gt_only}, closed-source models consistently outperform open-source alternatives. The best closed-source systems (Gemini-3-Image-Pro and GPT-Image-2) achieving similarity scores above 78, nearly doubling their predecessors' performance. However, high semantic similarity does not guarantee targeted scientific correction: while the latest models excel in visual reconstruction, their Question-ACC remains below 70. 
But earlier closed-source models with limited similarity scores (around or below 50) achieve Question-ACC exceeding 70. This discrepancy reveals that current frameworks prioritize global visual fidelity over localized knowledge integration, where similarity metrics may reflect pattern reproduction rather than accurate scientific content restoration.
Many open-source models exhibit low semantic consistency below 40, while maintaining moderate Question-ACC around 50, further underscoring the decoupling between visual reconstruction and scientific reasoning. Nevertheless, the latest models demonstrate partial usability for scientific applications, as their enhanced visual fidelity combined with moderate accuracy provides a foundation for refinement. These findings highlight the necessity for specialized generation frameworks that explicitly couple visual generation with scientific knowledge verification to achieve truly precise and controllable scientific image editing.

\paragraph{Upgradability Dependence on Correction Instruction Quality for Advanced Model}
\begin{wrapfigure}[21]{r}{0.48\textwidth}
    \centering
    \includegraphics[width=\linewidth]{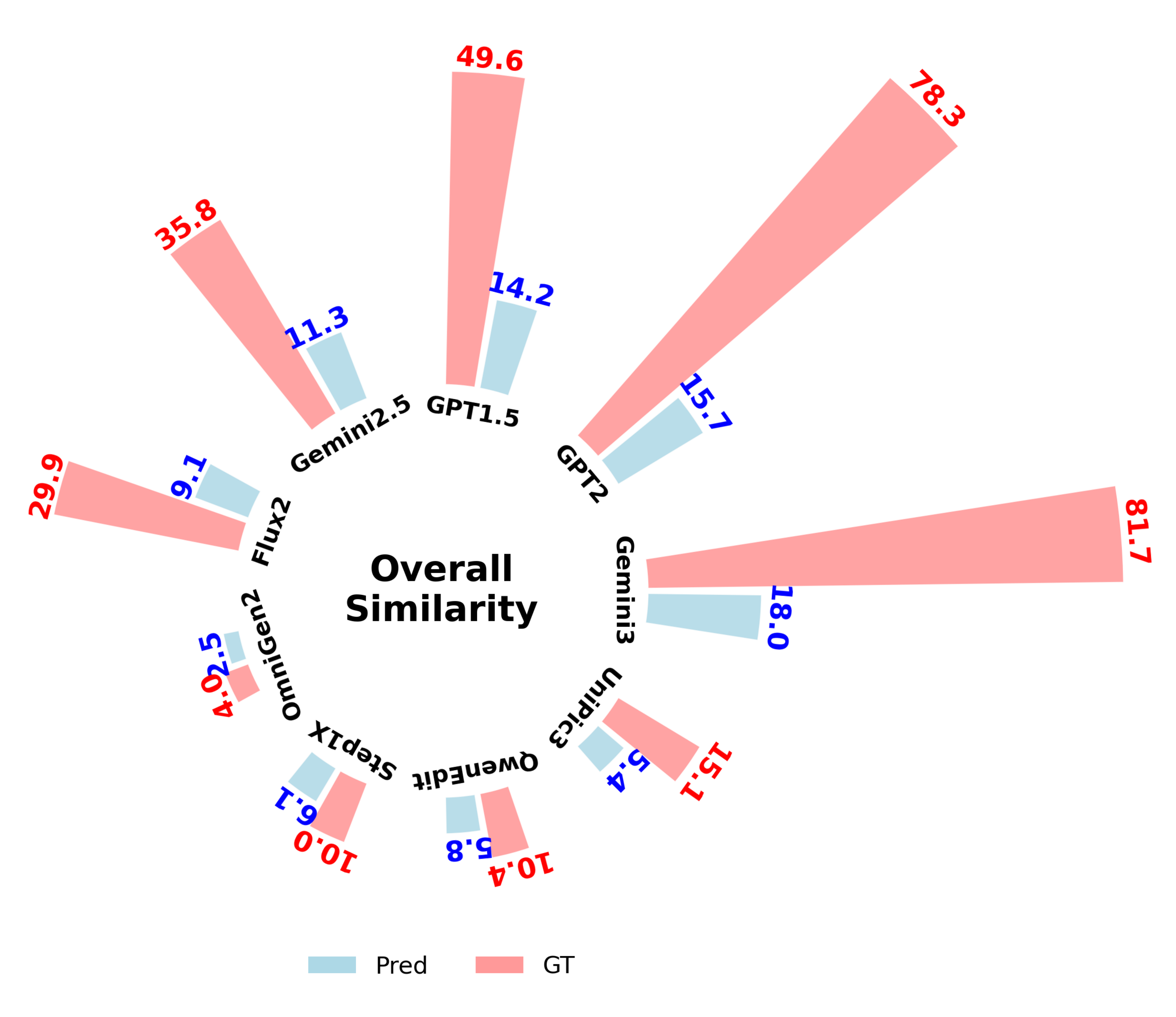} 
    \caption{Ablation study on upgradability: comparing ground-truth versus predicted correction instructions to assess their impact on editing performance.}
    \label{fig:ablation}
\end{wrapfigure}
The ablation results in Fig.~\ref{fig:ablation} reveal two distinct bottleneck regimes in how correction instruction quality impacts editing performance.
For high-performing models like Gemini-3-Image-Pro and GPT-Image-2, replacing predicted instructions (Pred) with human-annotated ground-truth instructions (GT) leads to dramatic improvements in semantic consistency, with Gemini-3-Image-Pro jumping from 18.1 to 81.6 and GPT-Image-2 surging from 15.7 to 78.6. This indicates that instruction quality serves as the primary bottleneck for these models, whose strong intrinsic generation capacity remains unrealized under imperfect guidance.
In contrast, low-performing models such as OmniGen2 exhibit only marginal improvements under GT instructions, climbing merely from 2.46 to 4.11. This suggests their performance is fundamentally constrained by limited editing capability rather than instruction quality. The divergence underscores a structural dependency in the editing pipeline: stronger models are bottlenecked by input instruction quality, while weaker models are bottlenecked by intrinsic generation capacity.

\section{Conclusion}


We introduce SIU$^2$A, a unified framework for scientific image reliability assessment through utility and upgradability, accompanied by SIU$^2$A-Benchmark for systematic evaluation. Our experiments reveal that current multimodal models struggle with grounded scientific error reasoning, and that restoration performance is fundamentally constrained by diagnostic quality. Through SIU$^2$A, we expose critical bottlenecks in existing approaches, highlighting the gap between error detection and correction capabilities. We believe this framework will serve as a foundational tool for developing more trustworthy AI in scientific applications, with potential extensions to domains requiring rigorous visual verification.

\bibliographystyle{abbrvnat}
\bibliography{sci-edit.bib}


\appendix
\section{Limitations}

This work introduces a novel evaluation framework (SIU2A), formulates a new task, and constructs a corresponding benchmark dataset. However, it does not propose dedicated methods to explicitly address or resolve the deficiencies identified through the evaluation (e.g., improving model robustness or editing capability). Instead, the primary goal is to systematically diagnose and quantify such limitations in current multimodal systems.

In addition, the effectiveness of the framework depends on the quality of generated instructions and error descriptions, which may introduce variability across different models or prompting strategies. While we attempt to standardize this process, residual noise may still affect evaluation outcomes.

Finally, although the benchmark covers diverse scientific image perturbations, it may not fully capture all real-world error distributions or domain-specific edge cases. We hope this work can serve as a foundation for future research on both more comprehensive benchmarks and targeted methods to address the identified limitations.

\section{Compute Resources}

All experiments were conducted using 4 $\times$ NVIDIA H200 GPUs. For proprietary large models, we accessed them via their official APIs under standard usage conditions. The computational cost mainly arises from large-scale evaluation across multiple models and benchmark instances. While individual inference calls are lightweight, the overall cost accumulates due to the multi-stage evaluation pipeline (diagnosis, instruction generation, and editing).

Based on our internal measurements under representative settings, the approximate API cost for standard utility evaluation using GPT-5.4 with GPT-5.2 as the evaluator is around \$12 per run. For upgradability evaluation, which involves image generation using GPT-Image-2 combined with evaluation using GPT-5.2 on advanced part, the cost is approximately \$40 per full evaluation run. These estimates may vary depending on API pricing, prompt length, and evaluation scale.
We note that the framework is designed to be modular and operates entirely in an evaluation-only setting without requiring any training of large models. However, reproducing the full benchmark results may still incur non-trivial compute or API costs due to large-scale inference.

\section{Open Access to Data and Code}

To support reproducibility and facilitate future research, the SIU\textsuperscript{2}A-Benchmark dataset and the evaluation toolkit are publicly available at the following locations.

\paragraph{Dataset.}
The dataset is hosted on Hugging Face:
\begin{center}
\url{https://huggingface.co/datasets/commusim-hf/SIUA}
\end{center}

\paragraph{Croissant Metadata.}
The dataset ships with a Croissant 1.1 metadata file augmented with
the Responsible AI (RAI) fields required by the NeurIPS 2026
Evaluations and Datasets Track, available at:
\begin{center}
\url{https://huggingface.co/datasets/commusim-hf/SIUA/resolve/main/croissant.json}
\end{center}
The minimal auto-generated Croissant record is also accessible via the
Hub API at
\url{https://huggingface.co/api/datasets/commusim-hf/SIUA/croissant}.

\paragraph{Evaluation Toolkit.}
The evaluation toolkit is available on GitHub:
\begin{center}
\url{https://github.com/commusim/SIUA-Eval}
\end{center}


\section{Prompt Design in the SIU$^2$A Framework}
\label{appendix:prompts}

The SIU$^2$A framework employs a three-stage evaluation pipeline to systematically assess scientific figure understanding and correction capabilities. This section provides the complete prompts used in each stage to ensure reproducibility of all experimental results.

\subsection{Utility 1: Error Detection}

The first stage focuses on identifying whether a scientific figure contains a visually identifiable error. The prompt instructs the model to act as a senior scientist reviewing the figure and requires precise localization of any detected error. The output format is strictly constrained to a JSON object to enable automated evaluation.

\promptbox{
You are a senior scientist reviewing a scientific figure. The figure may or may not contain a deliberate visual error.

Carefully examine the figure. If a specific and visually identifiable error exists (e.g., incorrect label, inconsistent trend, missing or swapped element), describe it precisely. The description must include both the location and the issue.

If no concrete and localizable error can be identified, indicate that no error is found.

Output \textbf{only} a valid JSON object in one of the following formats:

\begin{itemize}
\item Error detected: \\
\texttt{\{"found": true, "error\_description": "<location>: <issue>"\}}
\item No error: \\
\texttt{\{"found": false, "error\_description": null\}}
\end{itemize}
}

\subsection{Utility 2: Correction Feasibility}
Upon error detection, the second stage evaluates whether a definitive and unambiguous correction can be prescribed. This stage operationalizes the concept of correction feasibility by requiring the model to assess three critical conditions: precise error localization, high-confidence knowledge of the correct state, and the ability to express the correction as a single explicit instruction. The prompt filters out fundamentally indeterminate corrections that depend on unavailable information or have multiple plausible solutions.

\promptbox{
You are a scientific expert analyzing a figure that contains a known error.

Reported error: \texttt{\{error\_description\}}

Determine whether a \textbf{definitive and unambiguous correction} can be prescribed.

Output \texttt{\{"can\_instruct": true, "instruction": "<correction>"\}} if \textbf{all} of the following conditions are satisfied:

\begin{itemize}
\item The erroneous region can be precisely located in the figure
\item The correct state (value, label, structure, or trend) is known with high confidence, based on visual evidence or scientific knowledge
\item The correction can be expressed as a single, explicit, and unambiguous instruction specifying both the current error and the target state
\end{itemize}

Output \texttt{\{"can\_instruct": false, "instruction": null\}} if \textbf{any} of the following conditions hold:

\begin{itemize}
\item The error location cannot be identified
\item The correct state is fundamentally indeterminate (multiple plausible solutions exist)
\item The correct value depends on unavailable or author-specific information
\end{itemize}

Output \textbf{only} a valid JSON object.
}

\subsection{Upgradability: Image Restoration}

The final stage executes the actual image editing based on the instruction generated in Utility 2. The prompt is minimal by design, consisting solely of the repair instruction followed by a constraint that ensures localized modification. This constraint prevents unintended alterations to unrelated regions of the figure, preserving the overall scientific integrity of the visualization.

\promptbox{
\texttt{<instruction>}

\medskip
\textbf{Constraint:} Modify only the specified region. All other parts of the figure must remain unchanged.
}

\subsection{Pipeline Integration}

The three-stage design creates a cascading evaluation framework where each stage builds upon the output of the previous one. Stage 1 establishes the foundation by detecting concrete visual errors, Stage 2 ensures scientific rigor by filtering for well-defined corrections, and Stage 3 validates practical utility through precise image editing. This decomposition enables fine-grained analysis of model capabilities across perception, reasoning, and execution dimensions while maintaining strict control over error propagation throughout the pipeline.

\section{Data Construction Details}
\label{sec:appendix-data-construction}

Our dataset is constructed through a meticulously designed, three-stage pipeline that ensures the scientific validity, visual plausibility, and logical coherence of the generated perturbations. This process transforms high-quality base images into strictly paired (correct, erroneous) samples grounded in real-world scientific fallacies.

\paragraph{Stage a: High-Quality Base Image Selection.}
We begin by curating raw scientific figures from authoritative public sources, including ScienceQA, GenExam, BMMR, ChemVLM, and OmniScience. To guarantee both high visual fidelity and rich, editable content, we apply a two-step filtering protocol:
(1) \textit{Automated Quality Screening} using SIQA~\cite{li2026siqa}, a no-reference model trained on scientific imagery, to assess visual clarity, compositional structure, and information density; and
(2) \textit{Editability Rescreening} via an LLM to confirm the presence of identifiable, modifiable scientific elements (e.g., data series, labels, structural components).
This yields a pool of 600 high-fidelity base images as our ground-truth references.

\paragraph{Stage b: Perturbation Generation \& Curation.}

For each image, an LLM identifies ``Critical Visual Logic Anchors'' using this prompt:

\promptbox{
Role: Senior Scientific Image Forensic Expert.

Task: Identify 'Critical Visual Logic Anchors'—elements (Geometric Structure or Functional Text) defining the scientific logic.

Logical Coherence Gate (all must be YES):
\begin{itemize}
    \item Q1. FUNDAMENTAL: Is it a core logic carrier? (Not a generic label)
    \item Q2. CONTRADICTION: Does altering it create a fundamental scientific contradiction?
    \item Q3. RECOVERABILITY: Can an expert restore it from visual context alone?
\end{itemize}

Rules: Prioritize geometry over text. Text is valid only as a logic switch (e.g., 'D/A'). Exclude descriptive labels.

Output: JSON with 'context' and 'targets'. Each target has: 'category', 'description', 'location', 'original\_content', 'significance'.
}

This produces 4--8 anchors per image.

For each anchor and SIU$^2$A defect type (\textsc{Add}, \textsc{Modify}, \textsc{Remove}, \textsc{Swap}), we generate and curate perturbations.

\noindent\textbf{Plan Generation.}
An LLM creates candidate plans using this prompt:

\promptbox{
Context: \{context\}. Anchors: \{targets\_str\}.

Generate \{N\} plans for \{op\_type\} that create a scientifically illogical but visually plausible error.

Constraints:
\begin{itemize}
    \item Focus on geometric elements; functional text only if a logic switch.
    \item Error must be a LOGIC ERROR (narrative collapse), not just renaming.
    \item Must be DETECTABLE and RESTORABLE.
\end{itemize}

Output: JSON list. Keys include 'target\_description', 'location' (or 'location\_a/b'), 'surgical\_instruction', 'gt\_restoration\_instruction'. ONLY JSON.
}

\noindent\textbf{Plan Selection.}
The best plan is selected by an LLM evaluator using the FDRS criteria:

\promptbox{
Role: Scientific Image IUA Benchmark Constructor.

Task: Select the ONE best plan for '\{op\_type\}' that creates a Scientific Logical Error.

FDRS Criteria (ALL required):
\begin{itemize}
    \item \textbf{F: Feasible} (visually locatable/executable, no text rendering)
    \item \textbf{D: Detectable} (obvious, describable, scientific)
    \item \textbf{R: Restorable} (via visual context or domain knowledge)
    \item \textbf{S: Substantive} (alters core info, is misleading)
\end{itemize}

Output ONLY: an integer index (e.g., 0) or NONE.
}

The final pipeline generates ~2,100 validated image pairs.

\section{Annotation}
\label{sec:appendix-annotation}

To ensure the highest quality and scientific validity of our dataset, we implemented a rigorous expert-driven annotation pipeline. The process began by recruiting a panel of domain experts, including PhD candidates, postdoctoral researchers, and faculty members, with specialized knowledge in medicine, chemistry, biology, and physics.

Each expert was assigned figures strictly within their area of expertise to guarantee accurate and authoritative judgments. Prior to the main annotation phase, all experts participated in a comprehensive training session. This session covered the core error taxonomy (\textsc{Add}, \textsc{Modify}, \textsc{Remove}, \textsc{Swap}), provided detailed examples for each category, and established clear guidelines for writing precise, actionable repair instructions.

\begin{figure}[htbp]
    \centering
    \includegraphics[width=\linewidth]{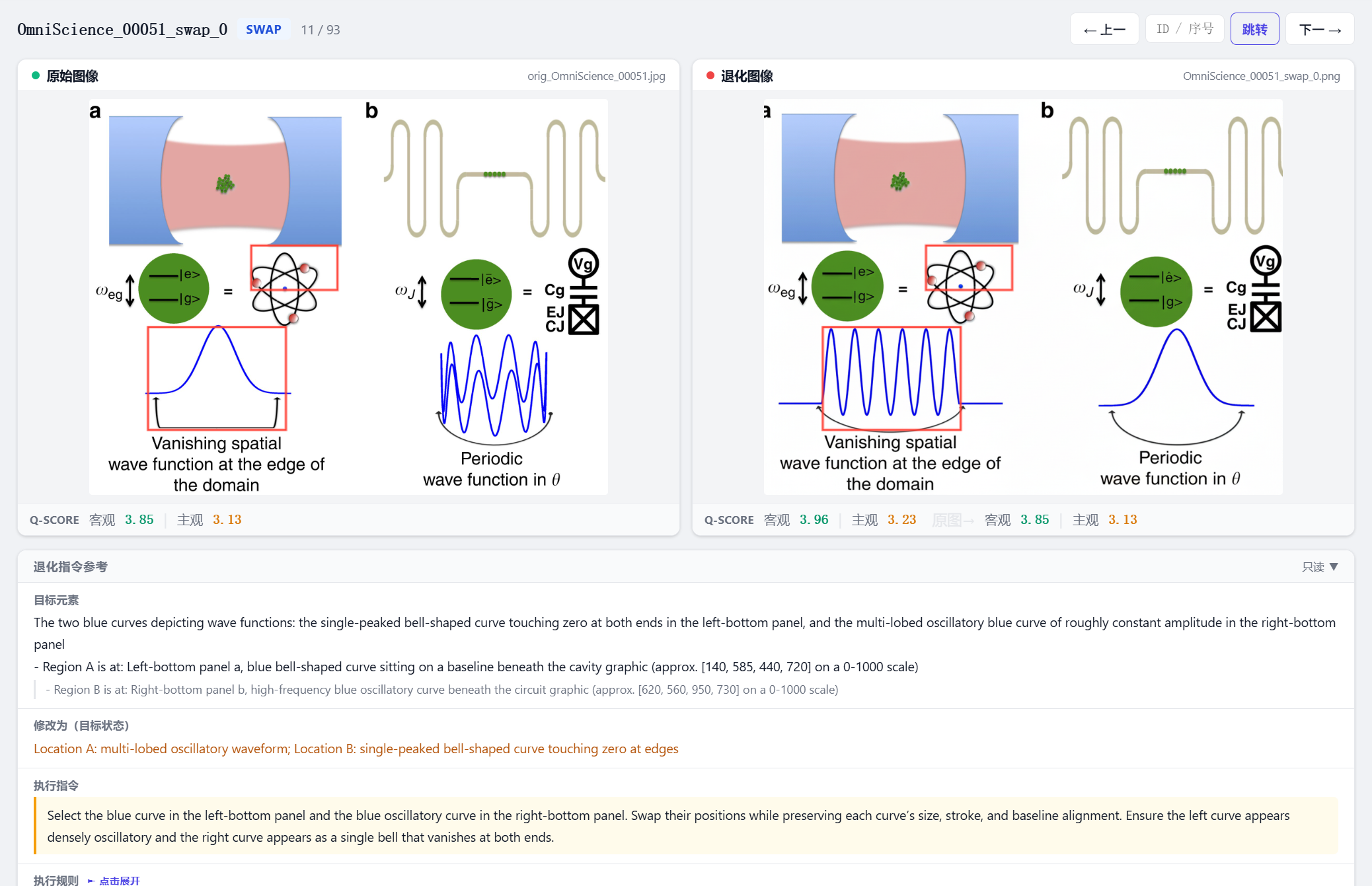}
    \includegraphics[width=1\linewidth]{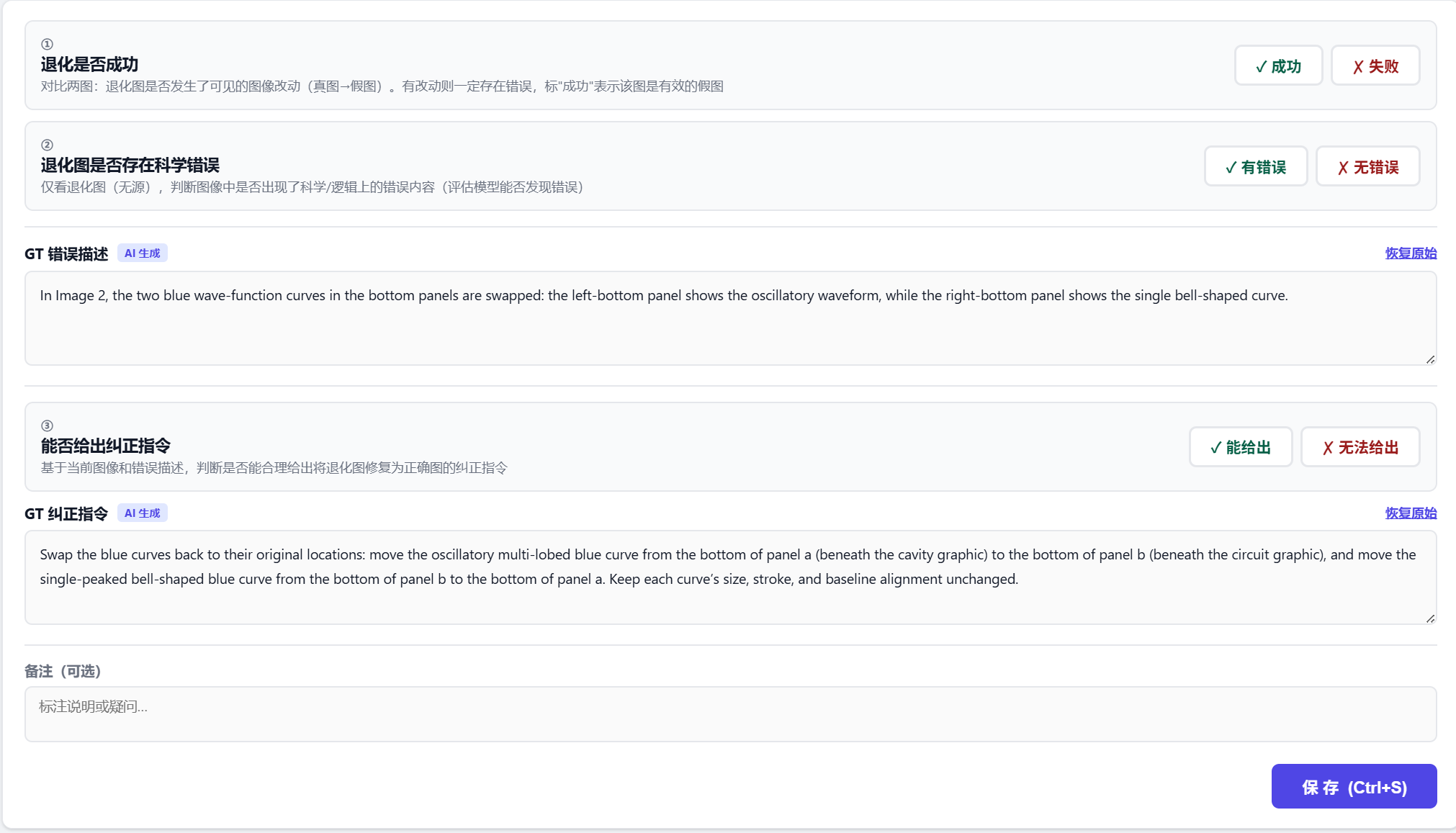}
    \caption{Our custom annotation interface. The tool presents the scientific figure to the expert annotator and provides dedicated fields for (1) describing the detected error with its location and (2) authoring a precise, executable instruction for its correction.}
    \label{fig:interfce}
\end{figure}

The annotation task was performed through a custom-built web interface, shown in Figure~\ref{fig:interfce}. The interface was designed to streamline the two-stage annotation process: first, experts identified and localized any visual errors in the figure, and second, they authored a concise, unambiguous instruction to correct the error. To maintain consistency and address edge cases, we established a continuous feedback loop. Experts were encouraged to engage in real-time discussions with the research team throughout the annotation process. This collaborative dialogue allowed us to clarify ambiguous scenarios, refine the annotation guidelines as needed, and ultimately achieve a high standard of expert-level annotations across the entire dataset.

\section{Appendix Case Study Visualizations}
\label{sec:appendix-case}

To complement the quantitative results presented in the main paper, this section provides a qualitative analysis through a curated set of case studies. We select four representative examples, each corresponding to one of the core error types (\textsc{Add}, \textsc{Modify}, \textsc{Remove}, \textsc{Swap}) and drawn from distinct advanced scientific domains (medicine, chemistry, biology, and physics). These same four cases are consistently used across all visualizations to enable end-to-end tracking of model behavior throughout the SIU$^2$A pipeline.

\subsection{Utility 1: Error Detection}
The visualizations are structured into a series of figures, each corresponding to a stage of our framework and designed with a consistent \emph{case-major} grid layout. In this layout, the four cases form the horizontal super-columns, while rows represent different data sections or model outputs.

\begin{figure}[!htbp]
  \centering
  \includegraphics[width=\textwidth]{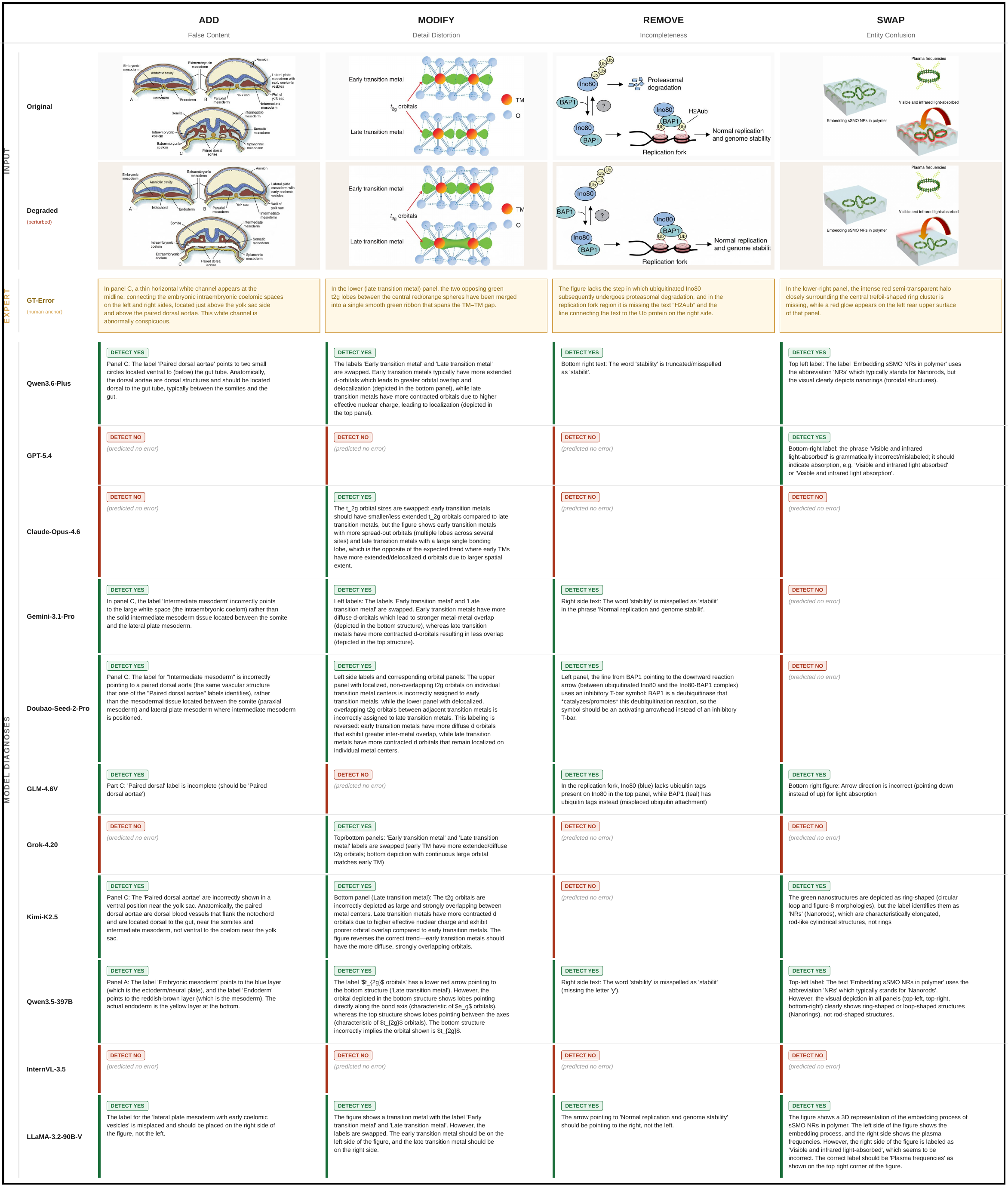}
  \caption{\textbf{Utility 1 --- Error Detection.} The figure displays the input images alongside the ground-truth expert annotations and the outputs from eleven diagnostic VLMs. Each model's prediction is encoded with a color bar (green for \emph{Detect: YES}, red for \emph{Detect: NO}), allowing for immediate assessment of detection accuracy against the known ground truth.}
  \label{fig:appendix-detect}
\end{figure}

\subsection{Utility 2: Correction Feasibility}

Building upon the detection results, the structured diagnosis stage is illustrated across two paginated figures. Each case super-column is split into two sub-columns to compare instruction generation under two conditions: \textbf{gold-conditioned} (using the human-authored ground-truth error description) and \textbf{pred-conditioned} (using the model's own predicted error). This direct comparison visually demonstrates how errors in the initial detection stage propagate into the quality and feasibility of the generated repair instructions.

\begin{figure}[!htbp]
  \centering
  \includegraphics[width=\textwidth]{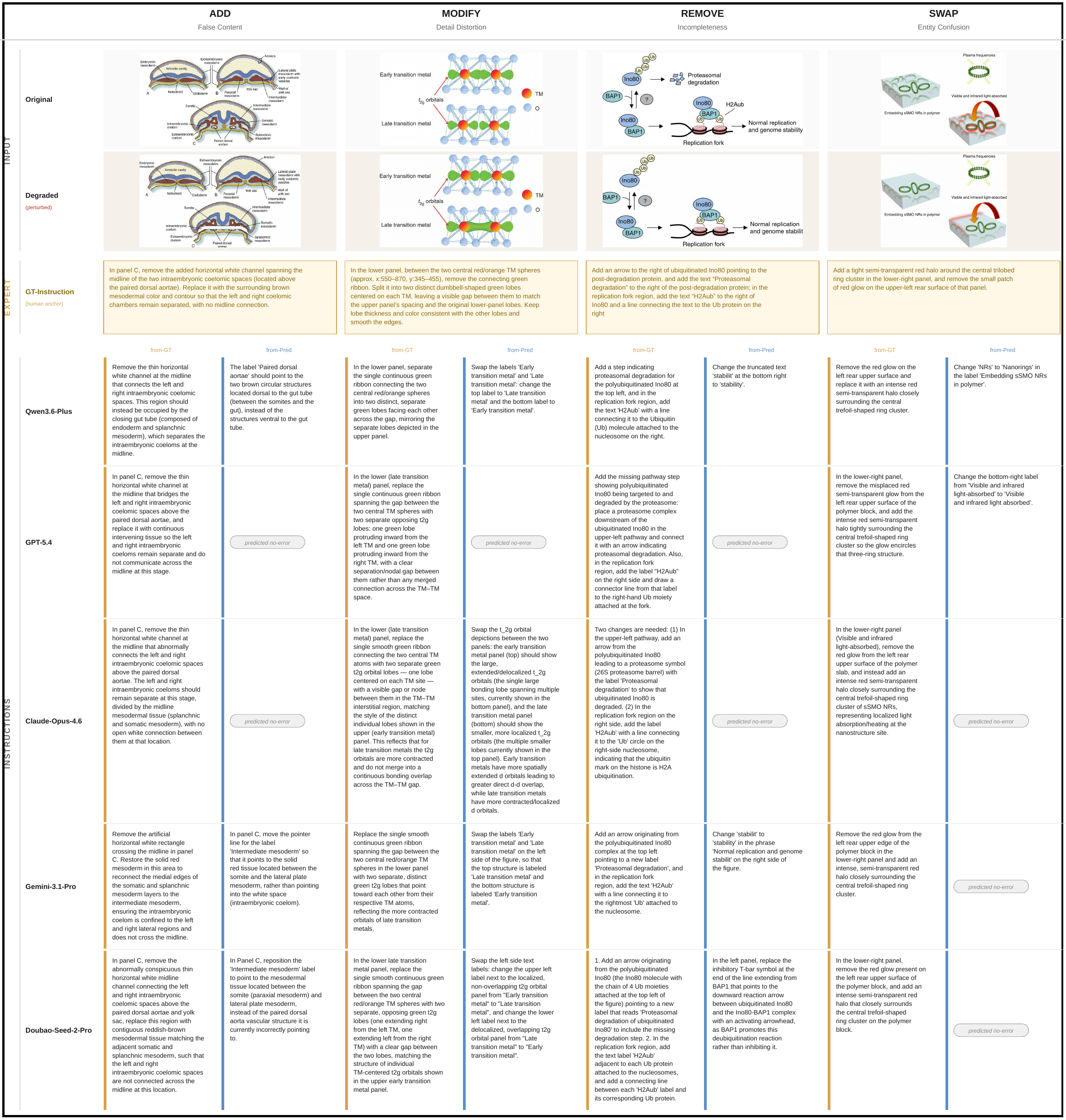}
  \caption{\textbf{Utility 2 --- Correction Feasiblity (page 1 of 2).} Results for the first five diagnostic models. The full-width gold \textsc{Expert} row reproduces the human GT-Instruction as a reference. A \emph{predicted no-error} chip replaces a missing instruction when a model declined to flag an error.}
  \label{fig:appendix-plan-p1}
\end{figure}

\begin{figure}[!htbp]
  \centering
  \includegraphics[width=\textwidth]{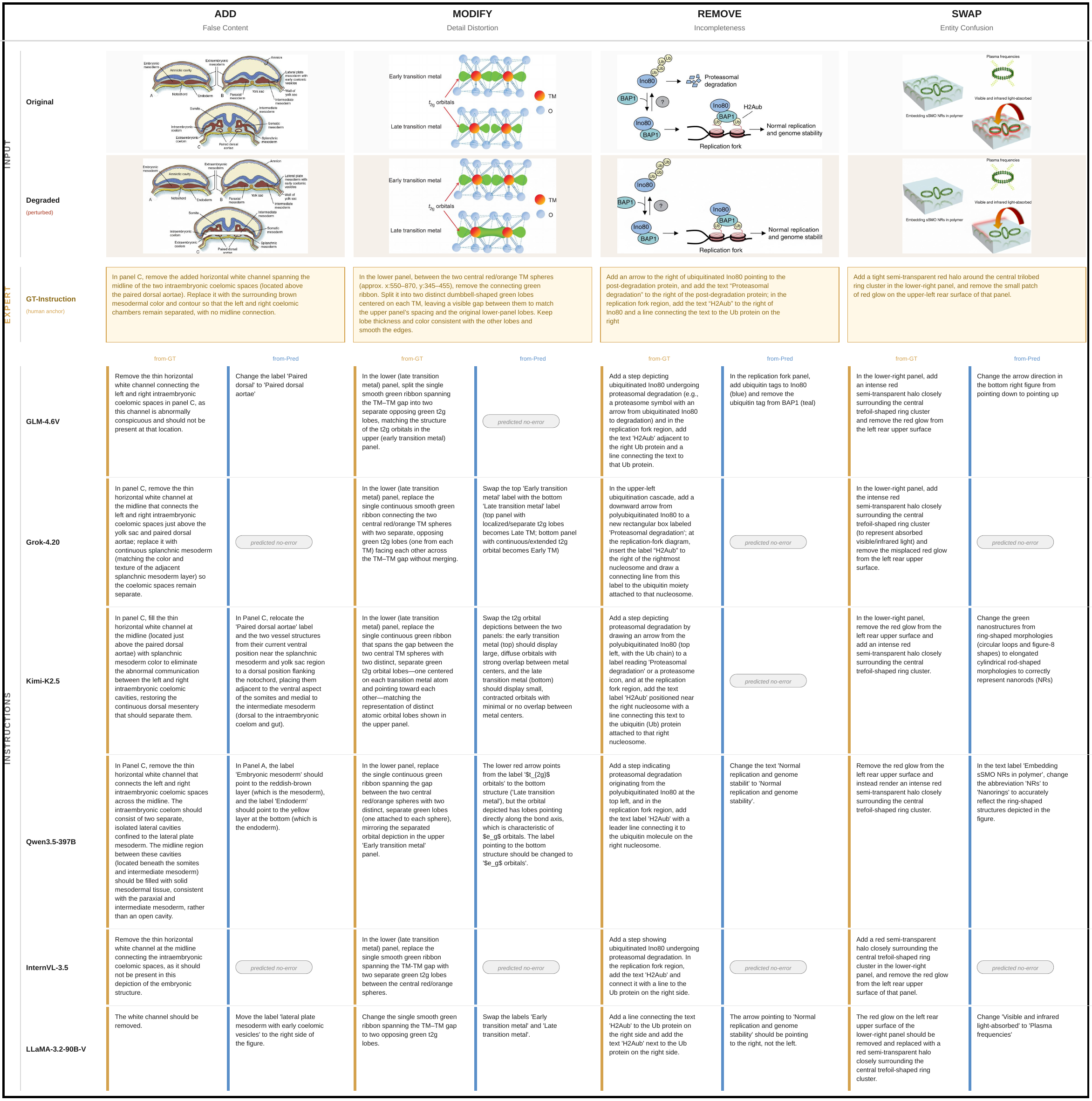}
  \caption{\textbf{Utility 2 --- Correction Feasiblity (page 2 of 2).} Results for the remaining six diagnostic models, with identical layout and semantics as Figure~\ref{fig:appendix-plan-p1}.}
  \label{fig:appendix-plan-p2}
\end{figure}

\clearpage

\subsection{Upgradability: Image Restoration}
Finally, the image restoration capabilities of Stage 3 are showcased in the following paginated figures. To isolate the performance of the editing models from upstream variance, the \textbf{pred-conditioned} sub-column uses a fixed repair instruction generated by the top-performing diagnostic model, Qwen3.6-Plus. The side-by-side comparison between \textbf{gold-conditioned} and \textbf{pred-conditioned} outputs provides a qualitative measure of the performance gap attributable solely to instruction quality.

\begin{figure}[!htbp]
  \centering
  \includegraphics[width=\textwidth]{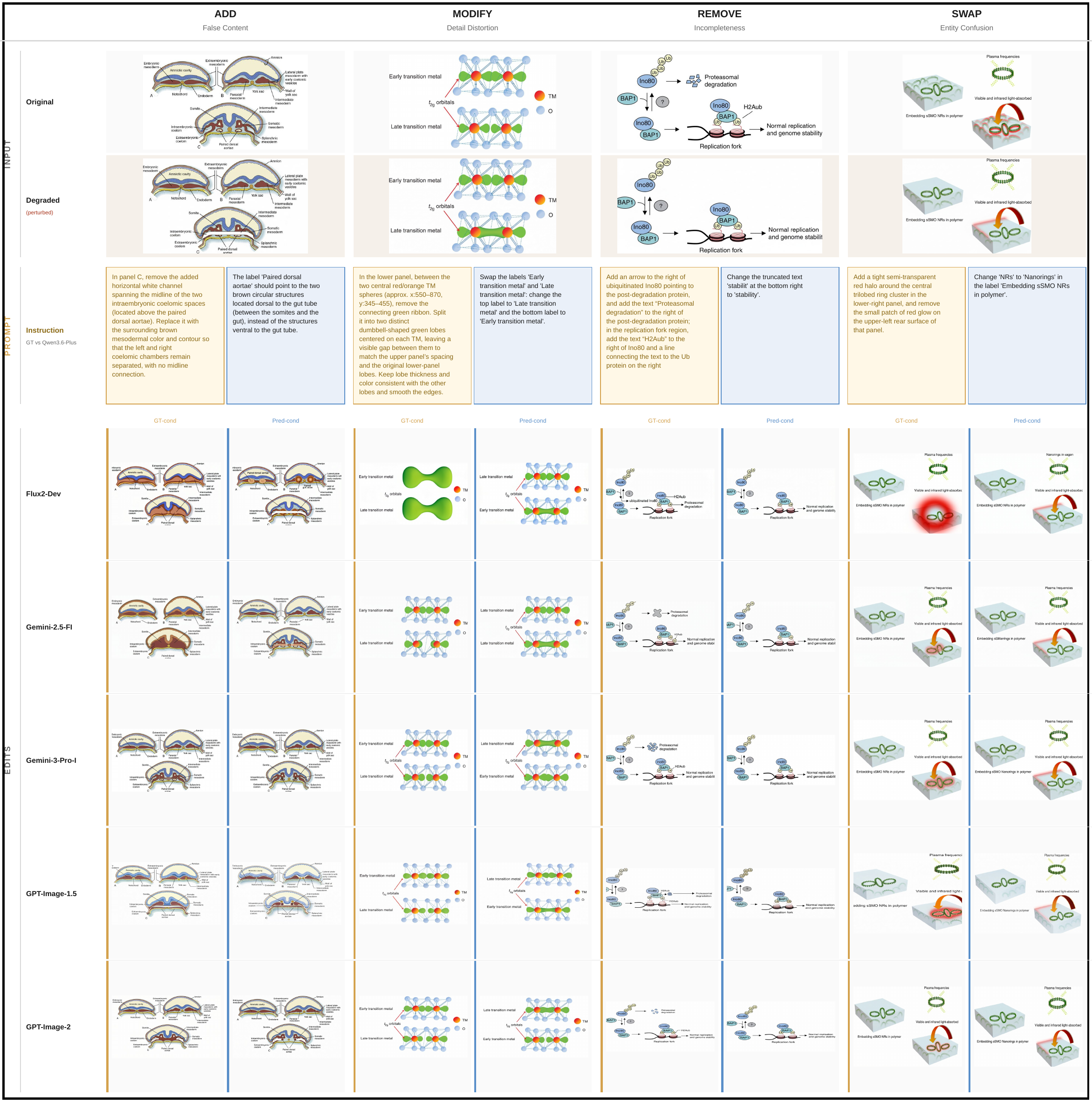}
  \caption{\textbf{Upgradability --- Image Restoration (page 1 of 2).} Results for the first five image editing models. The \textsc{Prompt} band surfaces both prompts side-by-side. The OmniGen-2 Pred-cond cell carries an \emph{instruction too long} tag, indicating its sensitivity to instruction length.}
  \label{fig:appendix-edit-p1}
\end{figure}

\begin{figure}[!htbp]
  \centering
  \includegraphics[width=\textwidth]{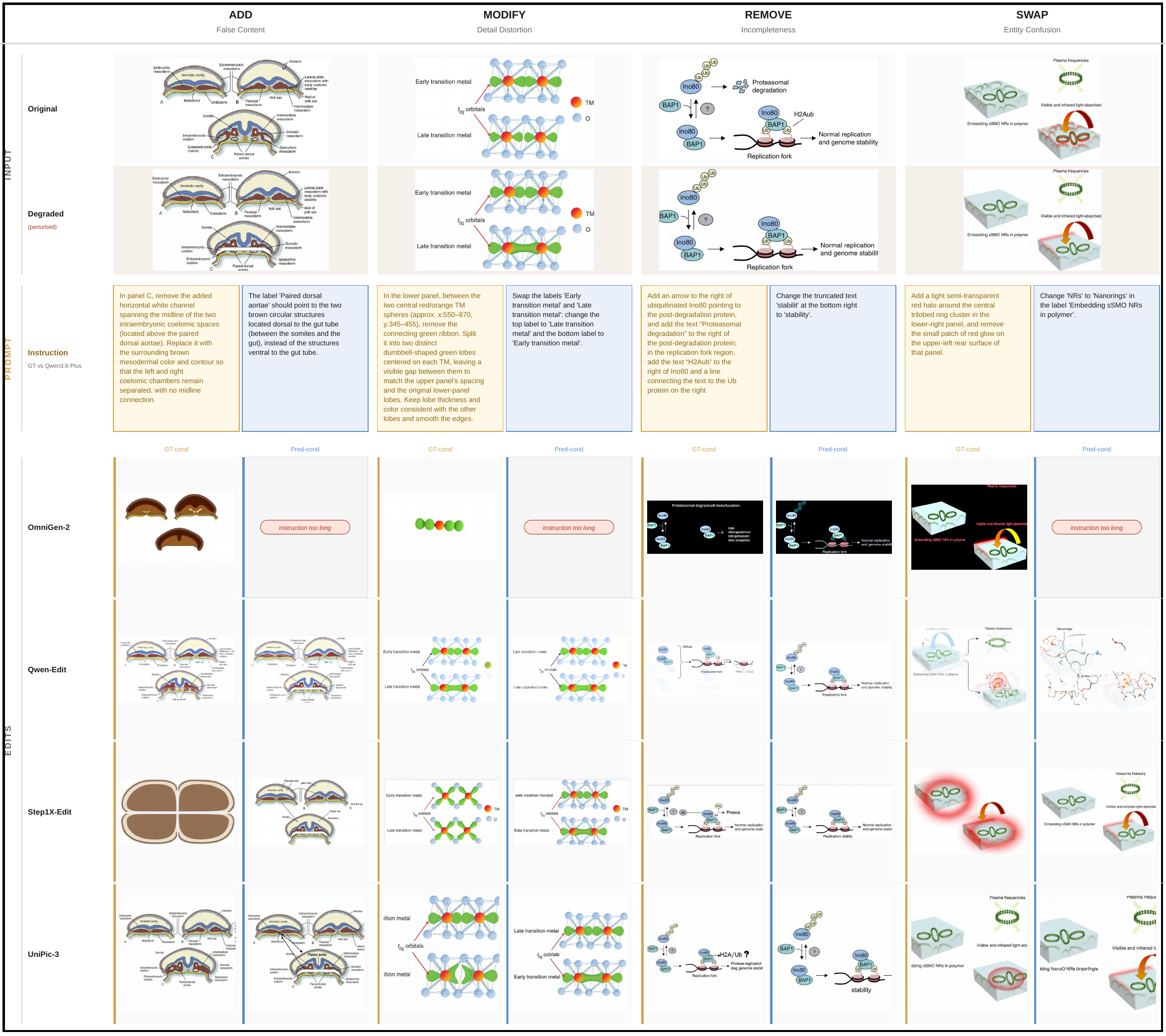}
  \caption{\textbf{Upgradability --- Image Restoration (page 2 of 2).} Results for the remaining four image editing models, with identical layout and semantics as Figure~\ref{fig:appendix-edit-p1}.}
  \label{fig:appendix-edit-p2}
\end{figure}

\section{Additional Ablations}
\label{sec:appendix-ablations}

This section presents two key ablation studies that dissect the performance of our SIU$^2$A framework. The first study isolates the impact of the instruction source on the image editing stage, while the second evaluates the intrinsic capability of diagnostic models to generate feasible repair instructions.

\subsection{Ablation on Instruction Source (Predicted vs. Ground Truth)}
\begin{table}[!htbp]
\centering
\caption{Ablation Study on Instruction Source (Predicted vs. Ground Truth)}
\label{tab:ablation_instruct_clean}

\begin{tabular}{
l
c
*{3}{>{\centering\arraybackslash}m{0.6cm}}
*{3}{>{\centering\arraybackslash}m{0.6cm}}
*{3}{>{\centering\arraybackslash}m{0.6cm}}
}

\hline
\textbf{Model} & \textbf{Src}
& \multicolumn{3}{c|}{\textbf{Pred True Rate}}
& \multicolumn{3}{c|}{\textbf{Accuracy}}
& \multicolumn{3}{c}{\textbf{Text Similarity}} \\
& 
& \textbf{Sim.} & \textbf{Adv.} & \textbf{All}
& \textbf{Sim.} & \textbf{Adv.} & \textbf{All}
& \textbf{Sim.} & \textbf{Adv.} & \textbf{All} \\

\hline
\multicolumn{11}{l}{\textit{Closed-source}} \\
\hline

\multirow{2}{*}{Qwen3.6-plus}
& Pred & 91.5 & 93.7 & 92.0 & 71.3 & 73.3 & 71.8 & 20.7 & 10.5 & 18.1 \\
& GT   & 94.9 & 86.7 & 92.9 & 73.7 & 69.5 & 72.7 & 61.4 & 53.0 & 59.3 \\

\multirow{2}{*}{Doubao-Seed-2.0-Pro}
& Pred & 85.7 & 80.5 & 84.4 & 68.2 & 66.4 & 67.7 & 23.8 & 9.8 & 20.3 \\
& GT   & 97.5 & 85.2 & 94.4 & 74.7 & 69.2 & 73.3 & 69.1 & 56.8 & 66.0 \\

\multirow{2}{*}{Gemini-3.1-Pro}
& Pred & 72.1 & 64.2 & 70.1 & 66.1 & 57.3 & 63.9 & 21.6 & 10.0 & 18.7 \\
& GT   & 88.7 & 70.0 & 84.0 & 72.4 & 69.0 & 71.5 & 59.4 & 49.3 & 56.9 \\

\multirow{2}{*}{Claude-Opus-4.6}
& Pred & 54.5 & 50.4 & 53.5 & 52.9 & 52.2 & 52.7 & 10.3 & 4.6 & 8.9 \\
& GT   & 98.4 & 86.2 & 95.3 & 74.1 & 69.7 & 73.0 & 73.6 & 57.4 & 69.6 \\

\multirow{2}{*}{GLM-4.6V}
& Pred & 46.8 & 38.6 & 44.7 & 48.7 & 42.5 & 47.2 & 7.2 & 1.3 & 5.7 \\
& GT   & 93.5 & 71.9 & 88.1 & 72.4 & 63.4 & 70.1 & 62.9 & 48.7 & 59.3 \\

\multirow{2}{*}{Grok-4.20}
& Pred & 44.2 & 45.4 & 44.5 & 47.6 & 48.4 & 47.8 & 13.4 & 4.4 & 11.1 \\
& GT   & 95.5 & 68.4 & 88.7 & 72.8 & 59.5 & 69.5 & 69.7 & 50.3 & 64.8 \\

\multirow{2}{*}{GPT-5.4}
& Pred & 35.0 & 26.1 & 32.7 & 43.7 & 36.9 & 42.0 & 7.3 & 2.4 & 6.1 \\
& GT   & 95.6 & 69.2 & 89.0 & 73.6 & 63.6 & 71.1 & 70.7 & 52.0 & 66.0 \\

\hline
\multicolumn{11}{l}{\textit{Open-source}} \\
\hline

\multirow{2}{*}{Qwen3.5-397B}
& Pred & 81.4 & 90.0 & 83.6 & 68.0 & 68.5 & 68.1 & 20.5 & 10.5 & 18.0 \\
& GT   & 92.2 & 88.2 & 91.2 & 72.4 & 69.7 & 71.7 & 62.1 & 54.4 & 60.2 \\

\multirow{2}{*}{Kimi-K2.5}
& Pred & 63.2 & 62.7 & 63.0 & 58.9 & 55.7 & 58.1 & 16.5 & 5.6 & 13.8 \\
& GT   & 93.5 & 68.2 & 87.2 & 73.0 & 60.6 & 69.9 & 66.2 & 46.5 & 61.3 \\

\multirow{2}{*}{LLaMa-3.2-90B-Vision}
& Pred & 85.6 & 79.0 & 84.0 & 67.7 & 63.9 & 66.8 & 5.4 & 1.8 & 4.5 \\
& GT   & 99.4 & 92.0 & 97.6 & 75.1 & 75.1 & 75.1 & 61.1 & 54.3 & 59.4 \\

\multirow{2}{*}{InternVL3.5-241B}
& Pred & 17.2 & 8.0 & 14.9 & 33.6 & 30.3 & 32.8 & 2.5 & 0.2 & 1.9 \\
& GT   & 85.9 & 81.5 & 84.8 & 71.7 & 69.0 & 71.0 & 57.6 & 51.9 & 56.2 \\

\hline
\end{tabular}

\end{table}

Table~\ref{tab:ablation_instruct_clean} investigates the critical dependency of the structured diagnosis stage (Stage 2) on the quality of the upstream error detection. For each diagnostic model, we compare its performance when conditioned on its own predicted error description (\textbf{Pred}) versus the human-authored ground-truth error (\textbf{GT}). The results are reported across three metrics: \textbf{Pred\_True\_Rate} (the rate at which the generated instruction is deemed correct and executable), \textbf{Binary\_Acc} (the accuracy of the binary decision on whether an instruction can be generated), and \textbf{Avg\_Text} (the average length of the generated instruction in tokens).

The data reveals a significant performance gap between the \textbf{Pred} and \textbf{GT} conditions, particularly for open-source models. This demonstrates that errors or ambiguities in the initial detection stage directly propagate into the instruction generation phase, leading to a substantial drop in the feasibility and correctness of the proposed repairs. Closed-source models like Qwen3.6-plus show greater robustness, maintaining a high Pred\_True\_Rate even with their own predictions, which underscores their superior diagnostic consistency.

\subsection{Edit-only Ablation on Advanced Instructions}
\begin{table}[!htbp]
\centering
\caption{Edit-only Ablation on Advanced Instructions}
\label{tab:edit_only_clean}

\setlength{\tabcolsep}{4pt}

\begin{tabular}{
l
c
*{5}{>{\centering\arraybackslash}m{0.6cm}}
*{5}{>{\centering\arraybackslash}m{0.6cm}}
}

\hline

\textbf{Model} & \textbf{Src}
& \multicolumn{5}{c}{\textbf{Similarity (Full)}}
& \multicolumn{5}{c}{\textbf{Question-ACC (None)}} \\

&
& \textbf{All} & \textbf{Add} & \textbf{Mod} & \textbf{Rem} & \textbf{Swap}
& \textbf{All} & \textbf{Add} & \textbf{Mod} & \textbf{Rem} & \textbf{Swap} \\

\hline
\multicolumn{12}{l}{\textit{Closed-source}} \\
\hline

\multirow{2}{*}{Gemini-3-Image-Pro}
&  Pred & 18.1 & 20.8 & 18.4 & 15.3 & 17.2 & 63.5 & 57.8 & 70.0 & 59.8 & 66.3 \\
&    GT   & 81.6 & 76.8 & 86.0 & 87.9 & 76.4 & 68.1 & 73.9 & 67.7 & 67.1 & 62.2 \\

\multirow{2}{*}{GPT-Image-2}
&  Pred & 15.7 & 15.8 & 12.8 & 12.6 & 21.7 & 60.8 & 57.8 & 65.0 & 55.6 & 64.4 \\
&    GT   & 78.6 & 80.2 & 82.5 & 83.5 & 66.7 & 66.3 & 63.2 & 67.6 & 69.6 & 65.2 \\

\multirow{2}{*}{GPT-Image-1.5}
&  Pred & 14.2 & 16.2 & 11.0 & 15.5 & 14.3 & 60.4 & 53.0 & 65.0 & 59.3 & 65.1 \\
&    GT   & 49.8 & 54.9 & 46.7 & 51.8 & 45.0 & 76.7 & 81.4 & 76.0 & 80.4 & 67.8 \\

\multirow{2}{*}{Gemini-2.5-Flash}
&  Pred & 11.4 & 13.5 & 11.1 & 9.25 & 11.3 & 60.4 & 57.3 & 62.4 & 55.4 & 66.7 \\
&    GT   & 35.7 & 36.8 & 29.5 & 42.2 & 34.8 & 70.6 & 70.8 & 70.2 & 75.8 & 65.6 \\

\hline
\multicolumn{12}{l}{\textit{Open-source}} \\
\hline

\multirow{2}{*}{Flux2-dev}
&  Pred & 9.14 & 10.8 & 8.83 & 6.33 & 10.2 & 59.4 & 55.3 & 61.4 & 56.6 & 64.4 \\
&    GT   & 30.2 & 39.4 & 26.0 & 27.2 & 26.5 & 63.4 & 64.6 & 65.4 & 66.3 & 56.7 \\

\multirow{2}{*}{OmniGen2}
&  Pred & 2.46 & 2.50 & 3.00 & 1.25 & 3.25 & 35.2 & 31.3 & 53.3 & 30.0 & 30.0 \\
&    GT   & 4.11 & 6.23 & 3.91 & 2.23 & 3.61 & 32.3 & 32.7 & 36.5 & 25.0 & 34.4 \\

\multirow{2}{*}{Step1X-Edit}
&  Pred & 6.11 & 6.01 & 5.20 & 5.88 & 7.53 & 52.9 & 49.5 & 56.4 & 48.2 & 57.5 \\
&    GT   & 10.2 & 12.8 & 11.1 & 7.41 & 8.67 & 59.7 & 65.5 & 58.7 & 48.9 & 64.4 \\

\multirow{2}{*}{Qwen-Edit}
&  Pred & 5.75 & 5.50 & 4.87 & 5.65 & 7.16 & 56.7 & 52.4 & 52.5 & 57.8 & 65.5 \\
&    GT   & 10.5 & 16.7 & 6.59 & 7.40 & 10.6 & 54.1 & 56.6 & 53.9 & 52.2 & 53.3 \\

\multirow{2}{*}{UniPic3}
&  Pred & 5.39 & 4.91 & 4.52 & 7.07 & 5.34 & 52.7 & 53.4 & 55.5 & 44.6 & 56.3 \\
&    GT   & 15.2 & 19.7 & 12.3 & 12.2 & 16.1 & 57.6 & 69.9 & 59.6 & 51.1 & 46.7 \\

\hline
\end{tabular}
\end{table}

Table~\ref{tab:edit_only_clean} isolates the performance of the image restoration stage (Stage 3) by controlling for the instruction source. We evaluate a suite of image editing models under two conditions: using the human ground-truth repair instruction (\textbf{GT}) and using a fixed prediction from the top-performing diagnostic model, Qwen3.6-plus (\textbf{Pred}). Performance is measured using \textbf{Similarity (Full)}, which quantifies the pixel-level fidelity of the edited image to the ground-truth corrected figure, and \textbf{Question-ACC (None)}, which assesses the semantic correctness of the edit through VQA.

The results highlight a stark contrast between closed-source and open-source editors. Leading closed-source models (e.g., Gemini-3-Image-Pro, GPT-Image-2) achieve high similarity scores under the \textbf{GT} condition, confirming their strong editing capabilities when provided with a perfect instruction. However, their performance degrades dramatically under the \textbf{Pred} condition, illustrating their sensitivity to instruction quality. In contrast, open-source models exhibit lower absolute performance in both conditions, with some models like OmniGen2 showing extreme sensitivity to instruction phrasing and length, as evidenced by its very low similarity scores.


\clearpage
\section*{NeurIPS Paper Checklist}

\begin{enumerate}

\item {\bf Claims}
    \item[] Question: Do the main claims made in the abstract and introduction accurately reflect the paper's contributions and scope?
    \item[] Answer: \answerYes{}
    \item[] Justification: The abstract and introduction clearly state the SIU2A framework, its two core dimensions (utility and upgradability), the benchmark construction, and experimental findings, which are consistent with the methodology and results presented in Sections 1 and 4.

\item {\bf Limitations}
    \item[] Question: Does the paper discuss the limitations of the work performed by the authors?
    \item[] Answer: \answerYes{}
    \item[] Justification: The limitations of the work, including the focus on evaluation rather than solution development, are explicitly discussed in the Appendix (Section: Limitations).

\item {\bf Theory assumptions and proofs}
    \item[] Question: For each theoretical result, does the paper provide the full set of assumptions and a complete (and correct) proof?
    \item[] Answer: \answerNA{}
    \item[] Justification: The paper does not present formal theoretical results or proofs; it focuses on framework design, dataset construction, and empirical evaluation.

\item {\bf Open access to data and code}
    \item[] Question: Does the paper provide open access to the data and code, with sufficient instructions to faithfully reproduce the main experimental results, as described in supplemental material?
    \item[] Answer: \answerYes{}
    \item[] Justification: The dataset and evaluation code is publicly released on Hugging Face and GitHub upon acceptance, with detailed instructions provided for reproducibility (see Appendix: Open Access to Data and Code).

\item {\bf Experimental setting/details}
    \item[] Question: Does the paper specify all the training and test details (e.g., data splits, hyperparameters, how they were chosen, type of optimizer) necessary to understand the results?
     \item[] Answer: \answerYes{}
    \item[] Justification: The experimental setup includes model lists, prompting format, evaluation metrics, and dataset construction details (Sections 3 and 4, Appendix A and B), sufficient to understand and replicate experiments.

\item {\bf Experiment statistical significance}
    \item[] Question: Does the paper report error bars suitably and correctly defined or other appropriate information about the statistical significance of the experiments?
    \item[] Answer: \answerNo{}
    \item[] Justification: Each evaluation run in our setting involves both image generation and text generation across multiple models, leading to high computational and API costs. As a result, we do not perform repeated runs for statistical reporting such as error bars or confidence intervals, and instead report results from single-run evaluations.

\item {\bf Experiments compute resources}
    \item[] Question: For each experiment, does the paper provide sufficient information on the computer resources (type of compute workers, memory, time of execution) needed to reproduce the experiments?
    \item[] Answer: \answerYes{}
    \item[] Justification: The paper specifies the compute resources (4×H200 GPUs) and API-based evaluation for proprietary models in the Appendix (Section: Compute Resources).

    \item[] Guidelines:

\item {\bf Code of ethics}
    \item[] Question: Does the research conducted in the paper conform, in every respect, with the NeurIPS Code of Ethics \url{https://neurips.cc/public/EthicsGuidelines}?
    \item[] Answer: \answerYes{}
    \item[] Justification: The research involves benchmark construction and model evaluation using publicly available datasets and models, with no identified ethical violations.

\item {\bf Broader impacts}
    \item[] Question: Does the paper discuss both potential positive societal impacts and negative societal impacts of the work performed?
    \item[] Answer: \answerYes{}
    \item[] Justification: This work contributes to improving the reliability and interpretability of multimodal systems in scientific domains by providing a structured evaluation framework for diagnosing and correcting image-level errors, which can benefit scientific research, education, and high-stakes applications. Potential negative impacts include over-reliance on automated evaluation signals, as well as possible misuse of editing capabilities to manipulate scientific images. We mitigate these risks by positioning the framework as a diagnostic and evaluation tool rather than a generative system, and by encouraging human oversight in downstream use.
   
\item {\bf Safeguards}
    \item[] Question: Does the paper describe safeguards that have been put in place for responsible release of data or models that have a high risk for misuse (e.g., pre-trained language models, image generators, or scraped datasets)?
     \item[] Answer: \answerNA{}
    \item[] Justification: The work focuses on evaluation and benchmarking; no high-risk generative model or sensitive dataset is released.

\item {\bf Licenses for existing assets}
    \item[] Question: Are the creators or original owners of assets (e.g., code, data, models), used in the paper, properly credited and are the license and terms of use explicitly mentioned and properly respected?
    \item[] Answer: \answerYes{}
    \item[] Justification: The paper exclusively uses publicly available datasets and models (e.g., ScienceQA, ChemVLM, OmniScience), all of which are properly cited. These resources are open-access or released for research use, and their usage complies with their respective terms.

\item {\bf New assets}
    \item[] Question: Are new assets introduced in the paper well documented and is the documentation provided alongside the assets?
    \item[] Answer: \answerYes{}
    \item[] Justification: The SIU2A-Benchmark dataset is thoroughly documented, including construction pipeline, annotation process, and structure (Section 3 and Appendix B/C).

\item {\bf Crowdsourcing and research with human subjects}
    \item[] Question: For crowdsourcing experiments and research with human subjects, does the paper include the full text of instructions given to participants and screenshots, if applicable, as well as details about compensation (if any)? 
    \item[] Answer: \answerYes{}
    \item[] Justification: Expert annotation is described in detail (Appendix C), including annotator qualifications and workflow.

\item {\bf Institutional review board (IRB) approvals}
    \item[] Question: Does the paper describe potential risks incurred by study participants, whether such risks were disclosed to the subjects, and whether Institutional Review Board (IRB) approvals (or an equivalent approval/review based on the requirements of your country or institution) were obtained?
    \item[] Answer: \answerYes{}
    \item[] Justification: The expert annotation process was conducted with informed consent from all annotators, and participation was voluntary. The study does not involve sensitive personal data or interventions, and all procedures comply with standard ethical guidelines for data annotation.

\item {\bf Declaration of LLM usage}
    \item[] Question: Does the paper describe LLM usage in core methods?
    \item[] Answer: \answerYes{}
    \item[] Justification: LLMs are central to dataset construction (anchor detection, perturbation planning), evaluation (LLM-as-a-judge), and prompting pipeline, and are explicitly described throughout Sections 3 and Appendix A/B.

\end{enumerate}

\end{document}